\documentclass[preprint,12pt,authoryear]{elsarticle}

\usepackage{amsmath,amssymb,amsfonts}
\usepackage{algorithm}
\usepackage{algorithmic}
\usepackage{graphicx}
\usepackage{booktabs}
\usepackage{multirow}
\usepackage{array}
\usepackage{enumitem}
\usepackage{tikz}
\usetikzlibrary{positioning,arrows.meta,calc}
\usepackage{tabularx}
\usepackage{xcolor}
\usepackage{geometry}
\usepackage{hyperref}
\usepackage{float}
\usepackage{placeins}
\usepackage{subcaption}
\usepackage{pgfplots}
\usepackage{pgfplotstable}
\usepackage{tikz}
\usepackage[T1]{fontenc}
\usepackage{helvet}
\usepackage{tikz}
\newcommand{\figfont}{\fontfamily{phv}\selectfont}
\newcommand{\fscale}{0.8}   
\usetikzlibrary{shapes.geometric, arrows.meta, positioning}
\pgfplotsset{compat=1.18}

\usepgfplotslibrary{groupplots}

\definecolor{moshogreen}{RGB}{0,180,0}
\definecolor{moshored}{RGB}{220,50,50}
\definecolor{moshoblue}{RGB}{70,130,210}
\definecolor{moshopurple}{RGB}{130,80,200}
\definecolor{moshoorange}{RGB}{220,160,60}

\pgfplotsset{
  moshobars/.style={
    ybar,
    bar width=18pt,
    enlarge x limits=0.15,
    ymajorgrids=true,
    grid style={line width=0.3pt, draw=gray!30},
    axis line style={draw=gray!60},
    tick style={draw=none},
    nodes near coords,
    nodes near coords align={vertical},
    every node near coord/.append style={font=\small},
    legend style={at={(0.5,-0.18)}, anchor=north, legend columns=-1,
                  draw=none, font=\small},
    xtick=data,
    x tick label style={font=\small, rotate=0},
    y tick label style={font=\small},
    title style={font=\small\bfseries},
    label style={font=\small},
  }
}

\journal{Swarm and Evolutionary Computation}

\begin{document}

\begin{frontmatter}

\title{MOSHO: A Native Discrete Shark-Inspired Metaheuristic for Hardware-Aware Neural Architecture Search}

\author[esi]{Nazim Abdeerhmane Aouni}
\author[esi]{Zakaria Soualah Mohammed}
\author[esi]{Maroua Ogab}
\author[esi]{Nour El Imane Elbar}
\author[esi]{Rayane Rahmat Errahmane Smara}
\author[esi]{Billel Moussous}
\author[esi]{Adriane Anis Khaled}
\author[esi]{Malika Bessedik}
\author[esi]{Fatima Benbouzid-Si Tayeb}
\author[ibm]{Hadjer Benmeziane}

\address[esi]{Higher National School of Computer Science (ESI), Algiers, Algeria\\
  \texttt{\{mn\_aouni, mz\_soualahmohammed, mm\_ogab, mn\_elbar, mr\_smara, mb\_moussous, ma\_khaled, m\_bessedik, f\_sitayeb\}@esi.dz}}

\address[ibm]{IBM Research\\
  \texttt{hadjer.benmeziane@ibm.com}}
\begin{abstract}
Hardware-aware neural architecture search (NAS) requires jointly optimizing predictive accuracy and device-specific latency over discrete search spaces. While multi-objective swarm optimizers show promise, continuous formulations such as the Multi-Objective White Shark Optimizer (MOWSO) can suffer from quantization noise when rounded to integer-encoded architecture vectors. To address this limitation, we propose MOSHO (Multi-Objective Shark Hunting Optimization), a native discrete metaheuristic that replaces continuous motion equations with discrete architecture-editing operators. MOSHO integrates adaptive operator scheduling, energy-regulated restarts, and deferred crowding-distance truncation to enhance exploration and search efficiency. On the merged NAS-Bench-201/HW-NAS-Bench benchmark across 18 contexts, MOSHO achieves a median normalized hypervolume of 0.9987 and IGD$^{+}$ of 0.0064, yielding Pareto fronts competitive with NSGA-II while reducing search runtime by 96.4\% (0.64\,s vs.\ 17.72\,s). These results demonstrate that tailored discrete search mechanisms can enable swarm optimizers to achieve high-quality multi-objective solution quality with substantially lower search cost in hardware-aware NAS.
\end{abstract}


\end{frontmatter}

\section{Introduction}
\label{sec:intro}

Deep learning has become a fundamental approach for image classification and visual recognition, with Convolutional Neural Networks (CNNs) achieving strong performance across a wide range of applications \citep{elsken2019survey}. The performance of a CNN depends not only on its learned parameters but also on architectural choices, including operation types, kernel sizes, channel widths, activation functions, and connectivity patterns. Manually designing these configurations is challenging because the number of possible combinations grows rapidly with the number of architectural decisions. Hyperparameter Optimization (HPO) addresses this challenge by automatically searching for effective training and model hyperparameters, typically while keeping the network architecture fixed. Neural Architecture Search (NAS) extends this optimization paradigm by treating the architecture itself as a search variable \citep{elsken2019survey,ren2021survey}. NAS therefore searches an architecture space $\mathcal{X}$ for configurations offering desirable predictive performance. Early NAS approaches based on reinforcement learning and evolutionary algorithms demonstrated the potential of automated architecture design but often required substantial computational resources \citep{zoph2017,real2019}. Subsequent weight-sharing approaches and tabular NAS benchmarks have made architecture search more practical and reproducible \citep{liu2019darts,dong2020nasbench201}.

For deployment on resource-constrained edge and Internet-of-Things (IoT) devices, however, predictive accuracy alone is insufficient. Hardware-Aware Neural Architecture Search (HW-NAS) therefore incorporates hardware-related objectives such as inference latency, memory consumption, and energy usage into architecture optimization. In this work, we focus on the accuracy--latency trade-off and formulate HW-NAS as the multi-objective problem
\begin{equation}
\min_{\mathbf{x} \in \mathcal{X}}
\mathbf{F}(\mathbf{x}) =
\left(
1 - \mathrm{Accuracy}(\mathbf{x}, \mathcal{D}_{\mathrm{val}}),
\mathrm{Latency}_d(\mathbf{x})
\right)^T,
\end{equation}
where $\mathbf{x}$ denotes a discrete architecture, $\mathcal{D}_{\mathrm{val}}$ is the validation set, and $\mathrm{Latency}_d(\mathbf{x})$ is its inference latency on target device $d$. The objective is to identify a Pareto set of architectures providing different accuracy--latency trade-offs.

Multi-objective NAS has been addressed using evolutionary algorithms, reinforcement learning, and swarm intelligence. Methods such as NSGA-II \citep{kumar2025} provide effective Pareto-based search, while swarm-based methods offer alternative exploration--exploitation mechanisms, with shark-inspired optimizers such as the White Shark Optimizer (WSO) and its multi-objective extension MOWSO providing promising search mechanisms \citep{braik2022,zheng2025}.However, adapting continuous shark-inspired optimization mechanisms to discrete architecture spaces remains a challenge, motivating the development of optimization strategies that operate directly on discrete architectures.

In this work, we introduce MOSHO (Multi-Objective Shark Hunting Optimization), a discrete shark-inspired metaheuristic for hardware-aware NAS. MOSHO operates directly on integer-encoded architectures using discrete architecture-editing operators. It further incorporates archive-guided search, adaptive operator scheduling, energy-regulated restarts, and deferred archive truncation to balance exploration and exploitation while limiting multi-objective search overhead.

The remainder of this paper is organized as follows. Section~\ref{sec:related} reviews related work on hardware-aware neural architecture search and multi-objective swarm optimization. Section~\ref{sec:method} presents MOSHO and its discrete search mechanisms. Section~\ref{sec:results} describes the experimental setup and results. Finally, Section~\ref{sec:conclusion} concludes the paper and outlines future research directions.

\section{Related Work}
\label{sec:related}

Multi-objective NAS can be viewed through two complementary components: an evaluation strategy, which assigns each candidate architecture its accuracy and hardware cost, and a search mechanism, the outer loop that proposes candidates and steers the population toward the accuracy--latency trade-off. Section~\ref{subsec:related_hwnas} reviews the former; Section~\ref{subsec:related_swarm} reviews the latter.

\subsection{Architecture Evaluation Methods for Hardware-Aware NAS}
\label{subsec:related_hwnas}

Early NAS approaches based on reinforcement learning \citep{zoph2017} and evolutionary search \citep{real2019} demonstrated that architecture design could be automated, but at a prohibitive computational cost, since every candidate architecture was trained from scratch. Three complementary directions have since emerged to make this evaluation tractable, trading off fidelity, generality, and cost differently.
The first relies on weight-sharing supernetworks, in which a single over-parameterized network embeds the entire search space and candidate architectures are extracted as subgraphs sharing a common set of trained weights, avoiding the need to train each candidate independently \citep{pham2018enas,liu2019darts}. Once-for-All \citep{cai2020ofa} extends this idea to hardware-aware deployment by decoupling supernet training from architecture search entirely, so that a specialized sub-network can be extracted for a given device and latency budget without additional training. This family amortizes training cost across many deployment scenarios but still requires training a single, carefully regularized supernet, and provides no explicit multi-objective search procedure for the accuracy--latency trade-off once sub-networks are available.

The second direction replaces training-based evaluation with a learned surrogate model that predicts an architecture's accuracy and/or hardware cost directly from its encoding. NAS-Bench-301 \citep{siems2021nasbench301} demonstrates that surrogate models can support efficient evaluation over the DARTS search space, which contains approximately $10^{18}$ architectures, using only a fraction of the architectures for training. BANANAS \citep{white2021bananas} further couples a path-based neural predictor with Bayesian optimization to guide the search itself. Surrogate models trade the exactness of ground-truth evaluation for scalability to arbitrarily large architecture spaces.

The third direction relies on exhaustive tabular benchmarks, where every architecture in a search space is evaluated once and stored in a lookup table, so optimizer runs query exact, pre-computed values instead of training or predicting them. NAS-Bench-201 \citep{dong2020nasbench201} and its hardware-oriented counterpart HW-NAS-Bench \citep{li2021hwnas}, used throughout this paper, sacrifice the scale reachable by supernets or surrogates but remove training stochasticity and approximation error entirely, isolating optimizer behavior as the sole source of variability, which is essential for the controlled comparison in Section~\ref{sec:results}.

All three directions address how an architecture is evaluated, not how the accuracy--latency trade-off space is explored once that evaluation is available: an outer-loop optimizer is still required to search the resulting objective landscape and maintain a representative Pareto front under a finite budget. This is the role played by MOSHO. 
\subsection{Optimization Methods for Neural Architecture Search}
\newcommand{\bx}[7]{
  \pgfmathsetmacro{\fs}{#6*\fscale}%
  \pgfmathsetmacro{\fl}{#6*\fscale+1}%
  \node[draw=#5, rounded corners=2pt, fill=white, align=center,
        font=\figfont\fontsize{\fs}{\fl}\selectfont\bfseries,
        inner sep=1pt, minimum width=#3cm, minimum height=#4cm,
        text width=#3cm-2pt] at (#1,#2)
        {\hyphenpenalty=10000 \exhyphenpenalty=10000 #7};}

\begin{figure}[H]
\centering
\resizebox{\linewidth}{!}{%
\begin{tikzpicture}
\draw[blue!70!black, rounded corners=8pt, fill=blue!4] (0,0) rectangle (11,10);
\node[font=\figfont\fontsize{10}{12}\selectfont\bfseries, text=blue!60!black]
  at (5.5,9.55) {Evolutionary Algorithms};

\bx{1.5}{8.4}{2.4}{1.4}{blue!70!black}{7}{DEHB\\(Evolutionary\\Hyperband)}
\bx{4.1}{8.4}{2.4}{1.4}{blue!70!black}{7}{NSGA-II}
\bx{6.7}{8.4}{2.4}{1.4}{blue!70!black}{7}{MOEA\\(New Search\\Strategy)}
\bx{9.3}{8.4}{2.4}{1.4}{blue!70!black}{6.5}{MOEA\\(Bi-population\\+ Uniform\\Sampling)}

\draw[orange!80!black, rounded corners=8pt, fill=orange!8] (0.2,2.3) rectangle (10.8,7.3);
\node[font=\figfont\fontsize{8}{9}\selectfont\bfseries, text=brown!70!black]
  at (5.5,6.95) {Swarm Intelligence};

\draw[red!80!black, dashed, rounded corners=6pt] (0.4,4.05) rectangle (6.1,6.5);
\node[font=\figfont\fontsize{6}{7}\selectfont\bfseries, text=red!80!black]
  at (3.25,6.2) {Shark-Based Methods};
\bx{1.4}{4.95}{1.75}{1.4}{red!80!black}{6}{White Shark\\Optimizer}
\bx{3.25}{4.95}{1.75}{1.4}{red!80!black}{6}{Multi-objective\\White Shark\\Optimizer}
\bx{5.1}{4.95}{1.75}{1.4}{red!80!black}{6}{Shark Smell\\Optimization}

\bx{7.15}{4.95}{1.9}{1.6}{orange!80!black}{6.5}{Particle Swarm\\Optimization}
\bx{9.3}{4.95}{1.9}{1.6}{orange!80!black}{6.5}{Multi-objective\\Particle Swarm\\Optimization}

\bx{1.3}{3.15}{1.95}{1.3}{orange!80!black}{5.5}{Variable-length\\Particle Swarm\\Opt.}
\bx{3.35}{3.15}{1.95}{1.3}{orange!80!black}{6}{HiveNAS\\(Bee-inspired)}
\bx{5.4}{3.15}{1.95}{1.3}{orange!80!black}{6}{Monarch\\Butterfly\\Optimization}
\bx{7.45}{3.15}{1.95}{1.3}{orange!80!black}{6}{Rank-Based\\Improved\\Firefly}
\bx{9.5}{3.15}{1.95}{1.3}{orange!80!black}{6}{Enhanced Swarm\\Metaheuristics}

\bx{1.5}{1.2}{2.4}{1.5}{blue!70!black}{7}{Pareto Dominance\\Novelty Search}
\bx{4.1}{1.2}{2.4}{1.5}{blue!70!black}{7}{Differential\\Evolution}
\bx{6.7}{1.2}{2.4}{1.5}{blue!70!black}{7}{Multi-obj.\\Genetic\\Programming}
\bx{9.3}{1.2}{2.4}{1.5}{blue!70!black}{6.5}{Pareto Evolutionary\\NAS\\(Dvolver-style)}

\draw[violet!70!black, rounded corners=8pt, fill=violet!5] (11.2,0) rectangle (16.2,10);
\node[font=\figfont\fontsize{10}{12}\selectfont\bfseries, text=violet!70!black, align=center]
  at (13.7,9.2) {Learning- and\\trajectory-based\\methods};
\bx{13.7}{6.9}{3.8}{1.4}{blue!70!black}{7}{Simulated Annealing\\(and variants)}
\bx{13.7}{4.9}{3.8}{1.4}{blue!70!black}{7}{Monte Carlo / NRPA}
\bx{13.7}{2.9}{3.8}{1.4}{blue!70!black}{7}{Trajectory-based\\Local Search}
\bx{13.7}{0.9}{3.8}{1.4}{blue!70!black}{7}{Repulsion-dynamics\\Optimization}
\end{tikzpicture}}
\caption{State-of-the-art landscape considered before focusing on shark-inspired optimization for hardware-aware multi-objective NAS}
\label{fig:swarm_landscape}
\end{figure}
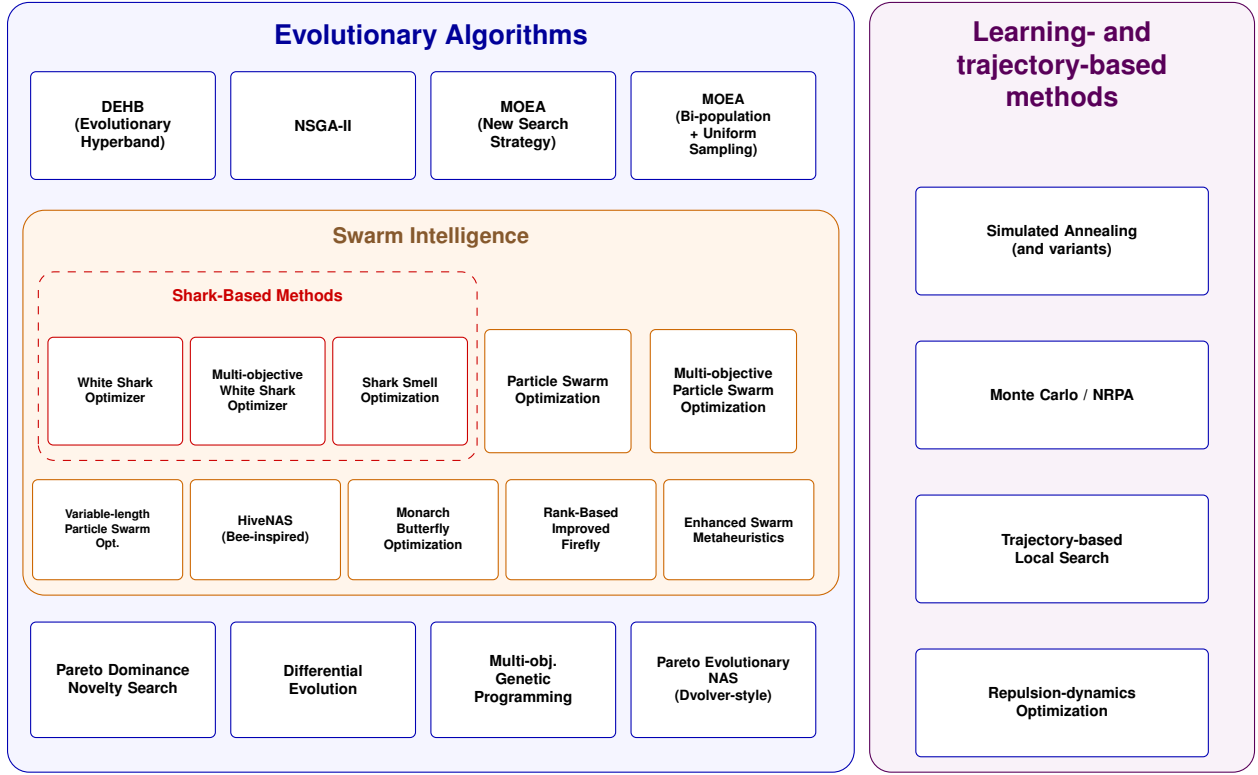

\label{subsec:related_swarm}

Multi-objective NAS employs a variety of search paradigms that can be broadly grouped into three categories: recombination-based evolutionary methods, swarm-intelligence methods, and Learning- and trajectory-based methods.

\textbf{Recombination-based evolutionary methods} encode architectures as discrete representations and evolve populations through selection and variation. Classical approaches such as NSGA-II\citep{kumar2025} use non-dominated sorting to maintain Pareto-optimal solutions, while decomposition-based methods optimize scalarized subproblems \citep{zhang2007}. Subsequent NAS variants have incorporated Bayesian-network exploitation \citep{lu2019nsganet}, co-evolutionary search with uniform-sampling initialization \citep{xue2026moeabus}, differential evolution combined with Hyperband \citep{awad2021dehb}, and alternative architecture encodings based on Cartesian genetic programming or trees \citep{garciagarcia2023cgp,michel2019dvolver}. Although these approaches provide effective population-based exploration, they often rely on relatively large populations and repeated objective evaluations, while adaptation to search history is generally limited.

\textbf{Learning- and trajectory-based methods} include reinforcement learning, Monte Carlo tree search, trajectory-based local search, and simulated-annealing-based approaches. Reinforcement-learning methods generate architectures as sequences of actions via policy gradients \citep{zoph2017}, while Monte Carlo tree search explores architecture spaces through simulated rollouts \citep{lallouet2025}. Trajectory-based methods can instead accept temporarily inferior candidates according to a decaying probability and may combine this mechanism with learned performance predictors \citep{liu2021foxnas}. These approaches provide alternative search mechanisms but can require substantial numbers of candidate evaluations, making their computational cost challenging under limited search budgets. Surrogate-assisted approaches, which combine learned performance models with an outer-loop search procedure, represent another class of methods and are discussed in Section~\ref{subsec:related_hwnas}.

\textbf{Swarm-intelligence methods} provide an alternative population-based paradigm based on cooperative or collective search behaviors. Multi-objective particle swarm optimization \citep{coello2004}, variable-length PSO \citep{wang2018pso}, artificial bee colony search \citep{shahawy2023}, firefly optimization \citep{nguyen2025}, monarch butterfly optimization \citep{bacanin2020mbo}, and Harris Hawks optimization \citep{basha2021hho} have been applied to NAS and related discrete optimization problems. However, most of these methods were originally formulated for continuous search spaces. Applying them to discrete architecture spaces therefore requires discretization or mapping mechanisms, which can introduce quantization effects and become less effective as architecture representations grow in complexity. Among swarm-intelligence approaches, shark-inspired methods are particularly relevant to this work:

\begin{itemize}
    \item \textbf{Shark-inspired methods.} These constitute the direct lineage of MOSHO. The White Shark Optimizer (WSO) \citep{braik2022}, its multi-objective extension MOWSO \citep{zheng2025}, and Shark Smell Optimization (SSO) \citep{kamarzaman2021} employ predator--prey and smell-guided mechanisms to balance exploration and exploitation. Detailed mathematical formulations of these methods are provided in \ref{app:supp_ancestors} of the Supplementary Materials. MOWSO extends WSO with a Pareto archive and True Distance-based leader selection but retains a continuous search formulation. Consequently, applying such methods to integer-encoded NAS architectures requires post-hoc rounding, which can introduce quantization effects \citep{zheng2025}. Moreover, maintaining an external Pareto archive with diversity-preserving truncation can introduce additional computational overhead as the population and archive grow. These limitations motivate the native discrete formulation and archive-management mechanisms introduced in MOSHO.
\end{itemize}
\section{Problem Formulation}
\label{subsec:formulation}

We address the joint optimization of the predictive accuracy and the hardware-specific inference latency of CNN architectures, a bi-objective discrete combinatorial problem whose solution is not a single architecture but a Pareto front of trade-offs \citep{hsu2018monas}.

\begin{figure}[htbp]
\centering
\includegraphics[width=0.45\linewidth]{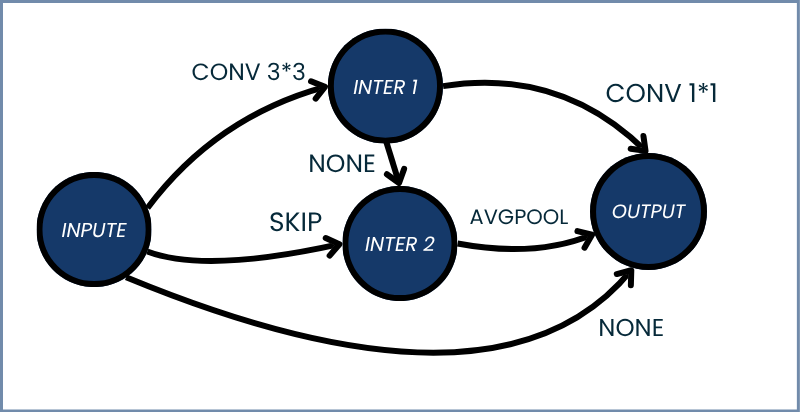}
\caption{Example NAS-Bench-201 cell (NONE: \texttt{zeroize}), encoded as $\mathbf{x}=(3,1,0,0,2,4)$ with indices $0$: \texttt{zeroize}, $1$: \texttt{skip-connect}, $2$: \texttt{conv}\,$1{\times}1$, $3$: \texttt{conv}\,$3{\times}3$, $4$: \texttt{avg-pool}\,$3{\times}3$.}
\label{fig:dag}
\end{figure}

\textbf{Search space.} We adopt the cell-based space of NAS-Bench-201 \citep{dong2020nasbench201}, also covered by HW-NAS-Bench \citep{li2021hwnas}, in which an architecture is entirely determined by a single cell stacked inside a fixed macro skeleton. The cell is a four-node directed acyclic graph whose $N_{\text{edges}}=6$ edges each carry one of $N_{\text{ops}}=5$ operations, $\{\texttt{zeroize},\ \texttt{skip-connect},\ \texttt{conv}\,1{\times}1,\ \texttt{conv}\,3{\times}3,\ \texttt{avg-pool}\,3{\times}3\}$, indexed $0,\ldots,4$ (Figure~\ref{fig:dag}). With the edge order $(0,1),(0,2),(1,2),(0,3),(1,3),(2,3)$, an architecture is the integer vector $\mathbf{x}=(x_1,\ldots,x_{N_{\text{edges}}})$, $x_j\in\mathcal{O}=\{0,\ldots,N_{\text{ops}}-1\}$, and $\mathcal{X}=\mathcal{O}^{N_{\text{edges}}}$ contains $5^6=15{,}625$ architectures.

\textbf{Objectives and Pareto optimality.} For a dataset $\mathcal{D}$ and a device $d$, each architecture has two conflicting objectives: the validation accuracy $f_1(\mathbf{x})=\mathrm{Accuracy}(\mathbf{x},\mathcal{D})$ (maximized) and the inference latency $f_2^{(d)}(\mathbf{x})=\mathrm{Latency}_d(\mathbf{x})$ (minimized), read from NAS-Bench-201 and HW-NAS-Bench, respectively. Architecture $\mathbf{a}$ \emph{dominates} $\mathbf{b}$, $\mathbf{a}\prec\mathbf{b}$, if $f_1(\mathbf{a})\geq f_1(\mathbf{b})$ and $f_2^{(d)}(\mathbf{a})\leq f_2^{(d)}(\mathbf{b})$, with at least one strict inequality \citep{miettinen1999, deb2002nsga2}. The true Pareto front $P^*=\{\mathbf{x}^*\in\mathcal{X} \mid \nexists\,\mathbf{x}\in\mathcal{X}:\mathbf{x}\prec\mathbf{x}^*\}$ is computed exactly for every dataset--device context by enumerating $\mathcal{X}$.

\textbf{Optimization goal.} Under a budget of $B$ evaluations, the optimizer seeks a set $\hat{P}$ of at most $M$ non-dominated architectures approximating $P^*$ as closely and uniformly as possible:
\begin{equation}
  \max_{\mathcal{E} \subseteq \mathcal{X},\ |\mathcal{E}| \leq B} \; I\bigl(\hat{P}, P^*\bigr),
  \qquad \hat{P} \subseteq \mathrm{ND}(\mathcal{E}),\quad |\hat{P}| \leq M,
  \label{eq:opt_goal}
\end{equation}
where $\mathcal{E}$ is the set of evaluated architectures, $\mathrm{ND}(\mathcal{E})$ its non-dominated subset, and $I$ a Pareto-compliant indicator such as the hypervolume (Section~\ref{sec:results}).

\textbf{Problem characteristics.} The space is categorical: operation indices carry no order, so velocity--position arithmetic has no architectural meaning, and the natural neighborhood is the Hamming distance between encodings \citep{awad2020, zheng2025}. The objectives are black-box \citep{dong2020nasbench201}, their trade-offs vary across devices \citep{li2021hwnas}, and the landscape is multi-modal \citep{elsken2019survey}, with plateaus created by functionally equivalent encodings (e.g., when \texttt{zeroize} disconnects an edge). These properties call for a natively discrete, archive-based metaheuristic.

\section{Proposed Method: MOSHO}
\label{sec:method}

The shark-inspired optimizers from which MOSHO descends were not designed for the problem of Section~\ref{subsec:formulation}. WSO \citep{braik2022} and SSO \citep{kamarzaman2021} are single-objective, and all three ancestors, including the multi-objective MOWSO \citep{zheng2025}, move agents through continuous velocity--position or gradient-following equations. On the categorical encoding of $\mathcal{X}$, these iterates must be rounded to integers: since operation indices have no order, a rounded step maps onto an arbitrary operation, and many continuous positions collapse onto the same architecture, so that the search trajectory is distorted by quantization noise. Moreover, the archive of MOWSO plays a passive role, supplying only the leader of the velocity update, while its control parameters follow fixed exponential schedules that ignore whether the search is progressing or stagnating. Finally, MOWSO prunes its archive with the True Distance metric after every insertion, a cost of $\mathcal{O}(N\cdot A^3)$ per iteration that dominates the total runtime when objective evaluations are cheap lookups. MOSHO addresses these limitations through three contributions: (i)~\textbf{six native discrete operator kernels} that edit integer architecture vectors directly, without relaxation or rounding (Section~\ref{subsec:operators}); (ii)~\textbf{an integrated adaptive search and archive-management strategy} combining energy-regulated restarts, a dual-signal operator scheduler, and deferred crowding-distance truncation to balance exploration and exploitation while reducing archive-maintenance overhead (Sections~\ref{subsec:energy}, \ref{subsec:scheduling}, and \ref{subsec:truncation}); and (iii)~\textbf{a comprehensive empirical evaluation} (Section~\ref{sec:results}) across 18 dataset--device contexts using the merged NAS-Bench-201/HW-NAS-Bench benchmark, assessing Pareto-front quality, search efficiency, and scalability against established baselines. The results show that MOSHO reduces search runtime by 96.4\% compared with NSGA-II while producing competitive Pareto fronts. A detailed structural comparison between MOSHO and its ancestors is provided in \ref{app:supp_comparison} of the Supplementary Materials.

\subsection{Architecture of MOSHO}
\label{subsec:components}

\begin{figure}[H]
\centering
\begin{tikzpicture}[
    scale=0.68, transform shape,
    core/.style={circle, draw=moshoblue!80, fill=moshoblue!10, thick, text width=1.9cm, align=center, font=\small\bfseries, minimum size=2.0cm},
    module/.style={rectangle, draw=gray!60, fill=gray!5, rounded corners=4pt, text width=3.0cm, align=center, font=\scriptsize, minimum height=0.8cm},
    active_module/.style={rectangle, draw=moshogreen!80, fill=moshogreen!5, rounded corners=4pt, text width=3.0cm, align=center, font=\scriptsize\bfseries, minimum height=0.8cm},
    arrow/.style={-{Latex[scale=0.8]}, thick, draw=gray!60},
    active_arrow/.style={-{Latex[scale=0.85]}, thick, draw=moshogreen!80, dashed}
]
    \node [core] (core) at (0,0) {MOSHO\\Core};

    \node [active_module] (pop)    at (90:3.3cm)  {Population \& Archive};
    \node [module]        (op)     at (18:3.3cm)  {Operator-Kernel Engine};
    \node [module]        (energy) at (306:3.3cm) {Energy Manager};
    \node [module]        (trunc)  at (234:3.3cm) {Deferred Truncation};
    \node [module]        (sched)  at (162:3.3cm) {Adaptive Scheduler};

    \path [arrow] (core) -- (pop);
    \path [arrow] (core) -- (op);
    \path [arrow] (core) -- (energy);
    \path [arrow] (core) -- (trunc);
    \path [arrow] (core) -- (sched);

    \path [active_arrow] (pop.south) to[out=-20, in=120]
        node[above right, font=\tiny, text=moshogreen!80!black,
        xshift=0.1cm, yshift=-0.15cm] {Gene sources} (op.north west);

    \path [active_arrow] (pop.east) .. controls (6.2, 2.6) and (5.4, -1.2) ..
        node[right, font=\tiny, text=moshogreen!80!black, xshift=0.05cm, yshift=0.4cm]
        {Recombination seeds} (energy.north east);

    \path [active_arrow] (pop.west) to[out=200, in=70]
        node[left, font=\tiny, text=moshogreen!80!black, xshift=-0.05cm]
        {Improvement signal} (sched.north);

    \path [arrow] (sched.east) --
        node[above, font=\tiny, yshift=0.05cm] {Operator prob.} (op.west);

\end{tikzpicture}
\caption{System architecture of MOSHO. Dashed green arrows denote the
archive-driven feedback loops that distinguish MOSHO from its ancestors.}
\label{fig:arch}
\end{figure}
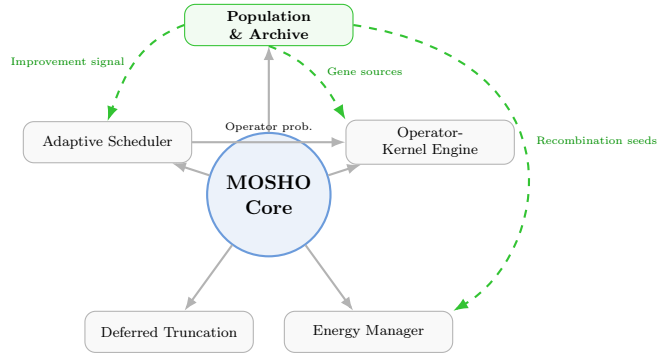

MOSHO evolves a population $\mathcal{P}^t$ of $N$ sharks, each characterized by an integer architecture vector $\mathbf{x}_i^t\in\mathcal{X}$ and an energy level $e_i^t$, together with an external Pareto archive $\mathcal{A}$ of bounded capacity $M$ that stores the best trade-offs found so far and constitutes the returned approximation set $\hat{P}$. Around this population and archive, the method is organized into four interacting modules: an \emph{operator-kernel engine}, which generates candidate architectures with six discrete operators sampled from a probability vector $\boldsymbol{\pi}^t$; an \emph{energy manager}, which tracks the success of each shark and restarts depleted ones; a \emph{deferred truncation} module, which bounds the archive while preserving its diversity; and a \emph{dual-signal adaptive scheduler}, which updates $\boldsymbol{\pi}^t$ from archive dynamics and search progress. As summarized in Figure~\ref{fig:arch}, the archive is not a passive store but the hub of the search: it supplies genes to the operators, seeds to the restarts, and an improvement signal to the scheduler.

\subsection{Discrete Hunting and Movement Operators}
\label{subsec:operators}

At each step, MOSHO samples an operator $o \sim \boldsymbol{\pi}^t$ and generates a candidate $\mathbf{x}' \sim K_o(\mathbf{x}_i^t, \mathcal{P}^t, \mathcal{A}^t)$. The six kernels replace the single velocity--position rule of WSO/MOWSO with complementary behaviors that differ in step size and in their source of genes: the agent itself, a population partner, the archive, or $\mathcal{O}$. All of them operate directly on integer vectors in $\mathcal{X}$ and require no rounding. Below, $\mathbf{x}=\mathbf{x}_i^t$ is the current position, $\mathbf{a}\in\mathcal{A}^t$ an archive member drawn with a bias toward large crowding distances (Section~\ref{subsec:truncation}), $\mathbf{z}$ a randomly selected partner from $\mathcal{P}^t$, and $u\sim\mathcal{U}(\mathcal{O})$ a uniformly random operation, drawn independently wherever it appears; genes not mentioned are copied from $\mathbf{x}$.

\begin{description}[itemsep=2pt, topsep=3pt, parsep=0pt, leftmargin=0pt]
\item[Patrol (long-range exploration).] A heavy-tailed multi-gene mutation, the discrete analogue of a L\'evy flight: $k$ distinct genes are replaced by random operations $u$, with
$k = \min\bigl(N_{\text{edges}},\, \max(2,\, \lfloor (v^{-0.35} \bmod N_{\text{edges}}) + 0.5 \rfloor)\bigr)$ and $v \sim \mathcal{U}(0,1)$. The heavy tail of $v^{-0.35}$ produces occasional large structural jumps that escape local optima and plateaus.

\item[Scent (odor-gradient following).] A gene-wise probabilistic takeover using the archive as a beacon:
\begin{equation}
  x'_j =
  \begin{cases}
    a_j & \text{with probability } 0.55, \\
    z_j & \text{with probability } 0.20, \\
    u   & \text{with probability } 0.05, \\
    x_j & \text{otherwise } (0.20),
  \end{cases}
  \qquad j = 1,\ldots,N_{\text{edges}}.
  \label{eq:scent}
\end{equation}

\item[Circle (close-range circling).] A local refinement in which $c\in\{1,2\}$ randomly selected genes receive a new operation $u \neq x_j$, so that $\mathbf{x}'$ lies at Hamming distance $c$ from $\mathbf{x}$.

\item[Burst (directed sprint).] A partial elite copy: $\mathrm{round}(\eta\, N_{\text{edges}})$ randomly chosen genes are copied from $\mathbf{a}$, with copy fraction $\eta = 0.33$ (two genes for $N_{\text{edges}} = 6$), and each remaining gene is replaced by $u$ with probability $0.35$.

\item[Crossover (recombination).] A uniform crossover with a partner, $\tilde{x}_j \in \{x_j, z_j\}$ with equal probability, after which, with probability $0.3$, one randomly selected gene of $\tilde{\mathbf{x}}$ is replaced by $u$.

\item[Scout (territory abandonment).] A full uniform reinitialization, $x'_j = u$ for all $j$, i.e., $\mathbf{x}' \sim \mathcal{U}(\mathcal{X})$.
\end{description}

Two properties of this set matter for NAS. Crossover has no analogue in continuous velocity--position dynamics, yet it is what recombines building blocks discovered along independent trajectories. Scent and Burst are the operators through which the archive directly steers movement, the discrete counterpart of following an odor gradient, whereas Patrol and Scout supply the diversification needed to leave plateaus.

\subsection{Energy-Regulated Restarts}
\label{subsec:energy}

Each shark carries an energy level $e_i \in [0,e_{\max}]$, initialized at $e_0 = 0.818$. Energy is updated at every step according to the outcome of the current search move:
\begin{equation}
  e_i^{t+1} =
  \begin{cases}
    \min(e_{\max},\, e_i^t + \beta_1) & \text{if the move improves the archive,} \\
    \min(e_{\max},\, e_i^t + \beta_2) & \text{if the move improves the current solution,} \\
    \max(0,\, e_i^t - \delta)         & \text{if the move is non-improving,}
  \end{cases}
  \label{eq:energy}
\end{equation}
where $\beta_1 = 0.20$, $\beta_2 = 0.05$, and $\delta = 0.0968$. A move improves the current solution when the candidate Pareto-dominates the shark's current position, in which case the shark moves to the candidate; a move that improves the archive is also always accepted.

An agent whose energy falls below $e_{\min} = 0.5$ is restarted using either \emph{archive recombination}, which performs uniform crossover between two randomly selected archive members followed by a $15\%$ random gene perturbation, or \emph{scout restart}, which fully reinitializes the architecture. The restart source is selected stochastically, and the energy is reset to $\max(\gamma\,e_0,\,e_{\min})$ with $\gamma = 0.5$. Energy thus converts the recent success of each shark into a budget of tolerated failures: productive sharks keep exploiting their region, whereas sharks trapped on a plateau are relocated, preferably to recombinations of elite solutions.

\subsection{Archive Management and Selection}
\label{subsec:truncation}

The archive is maintained under a dominance-first policy. Initially, $\mathcal{A}^0$ is built from the non-dominated subset of the uniformly sampled initial population. Afterwards, a candidate is inserted whenever no archive member dominates it, and all archive members it dominates are removed at insertion time, whereas truncation to the capacity $M$ is deferred to the end of the iteration. This decoupling of insertion from truncation is where MOSHO obtains most of its runtime advantage. MOWSO recomputes TD-based pruning after \emph{each} candidate insertion, which costs $\mathcal{O}(A^2)$ pairwise distances per candidate and $\mathcal{O}(N \cdot A^3)$ per iteration once the iterative removals are accounted for, where $A=|\mathcal{A}|$. MOSHO instead performs two distinct steps:

\begin{enumerate}
  \item \textbf{Per candidate, inside the agent loop:} a dominance check only, $\mathcal{O}(A)$. Non-dominated candidates are inserted without any sorting, and dominated members are dropped on the spot.
  \item \textbf{Per iteration, at the end of the loop:} if $|\mathcal{A}| > M$, a single crowding-distance-based sorting, $\mathcal{O}(A\log A)$, ranks archive members by local density and prunes the most crowded ones until $|\mathcal{A}| = M$. The extreme solutions of each objective are excluded from this pruning, so the boundaries of the front are never lost.
\end{enumerate}

The per-iteration cost therefore drops from $\mathcal{O}(N \cdot A^3)$ to $\mathcal{O}(N \cdot A + A\log A)$, roughly a factor $A^2$, and this single policy is the main source of the runtime gap reported in Section~\ref{sec:results}. Crowding distance is also the more suitable diversity metric here, since it measures per-axis spacing and preserves extreme solutions by construction, which TD does not. Leader selection follows the same principle: when Scent or Burst requires an archive member, the draw is biased toward large crowding distances, so movement is guided toward the sparse regions of the front rather than its crowded center.

\subsection{Dual-Signal Adaptive Scheduling}
\label{subsec:scheduling}

The operator probability vector is not fixed but modulated by two complementary feedback signals. The \emph{archive improvement rate} $r^t = \frac{1}{H}\sum_{h=t-H+1}^{t}\mathbf{1}[\Delta\mathcal{A}_h > 0]$ measures, over a sliding window of $H = 5$ iterations, the fraction of recent iterations in which the archive was improved. The \emph{search progress} $p^t = \min(1, \text{eval\_count}/B)$ measures the consumed fraction of the evaluation budget. Both signals are combined multiplicatively on the base probabilities $\boldsymbol{\pi}^0$:
\begin{equation}
  \boldsymbol{\pi}^t = \mathrm{Normalize}\bigl(\boldsymbol{\pi}^0 \odot g(r^t, p^t)\bigr),
  \label{eq:pi_update}
\end{equation}
where $g(r^t,p^t)$ is an operator-specific feedback function whose complete piecewise definition is provided in \ref{app:supp_extensions} of the Supplementary Materials. Its effect is as follows: early in the search ($p^t \to 0$), the long-range operators Patrol and Scout are boosted; as the budget is consumed ($p^t \to 1$), the weight shifts toward the archive-guided Scent and toward Crossover, which recombines the building blocks accumulated by the population; stagnation ($r^t \to 0$) further boosts Scout and Burst, while sustained archive improvement ($r^t \to 1$) favors the local refinement of Circle.

\subsection{Algorithm and Workflow}
\label{subsec:algorithm}

\begin{figure}[H]
\centering
\begin{tikzpicture}[
scale=0.58, transform shape,
node distance=0.7cm and 0.7cm,
block/.style={
rectangle, draw, fill=blue!5,
text width=6.2cm, align=center,
rounded corners, minimum height=0.7cm,
font=\small
},
subblock/.style={
rectangle, draw, fill=blue!2,
text width=4.0cm, align=center,
rounded corners, minimum height=0.7cm,
font=\small
},
decision/.style={
diamond, draw, fill=orange!10,
text width=2.0cm, align=center,
aspect=2, font=\small, inner sep=0.5pt
},
line/.style={
draw, -{Latex[scale=0.8]}, thick
},
drawline/.style={
draw, thick
},
umlstart/.style={
circle, fill=black,
minimum size=0.35cm, inner sep=0pt
},
umlfinal/.style={
circle, draw, thick,
minimum size=0.45cm, inner sep=0pt,
path picture={
\fill[black]
(path picture bounding box.center) circle (0.12cm);
}
}
]

\node [umlstart] (start) {};
\node [block, below=0.5cm of start] (init)
    {Initialize population $\mathcal{P}_0$\\
     Evaluate and build archive $\mathcal{A}_0$};

\node [block, below=of init] (progress)
    {Compute progress $p_t$, improvement $r_t$\\
     Update operator probabilities $\boldsymbol{\pi}_t$};

\node [block, below=of progress] (candidate)
    {Generate candidate $\mathbf{x}'_i$\\
     $o \sim \boldsymbol{\pi}^t$, $\mathbf{x}' \leftarrow
     K_o(\mathbf{x}_i^t,\mathcal{P}^t,\mathcal{A}^t)$};

\node [block, below=of candidate] (eval)
    {Lookup evaluation $\mathbf{y}'_i = E(\mathbf{x}'_i)$};

\node [block, below=of eval] (update)
    {Dominance-based archive update\\
     $\mathbf{x}_i^t \leftarrow \mathbf{x}'_i$};

\node [block, below=of update] (energy_update)
    {Energy update\\
     $+\beta_1$ if archive improved, $+\beta_2$ if individual improved,\\
     else $-\delta$};

\node [decision, below=of energy_update] (energy)
    {Energy $< e_{\min}$?};

\node [subblock, left=1cm of energy, yshift=-1.0cm] (yes_energy)
    {Archive recombination\\or scout restart};

\node [subblock, right=1cm of energy, yshift=-1.0cm] (no_energy)
    {Keep current shark state};

\node [decision, below=2.6cm of energy] (all_sharks)
    {All sharks\\processed?};

\coordinate [above=0.5cm of all_sharks] (merge2);

\node [subblock, right=1cm of all_sharks, yshift=-1.0cm] (yes_all)
    {Batch archive\\truncation};

\node [decision, below=of yes_all] (budget)
    {Budget\\exhausted?};

\node [subblock, below=of budget] (return)
    {Return Pareto\\front approximation};

\node [umlfinal, below=0.5cm of return] (end) {};

\path [line] (start) -- (init);
\path [line] (init) -- (progress);
\path [line] (progress) -- (candidate);
\path [line] (candidate) -- (eval);
\path [line] (eval) -- (update);
\path [line] (update) -- (energy_update);
\path [line] (energy_update) -- (energy);

\path [line] (energy.west)
    -| node[above, pos=0.3] {Yes} (yes_energy.north);

\path [line] (energy.east)
    -| node[above, pos=0.3] {No} (no_energy.north);

\path [drawline] (yes_energy.south) |- (merge2);
\path [drawline] (no_energy.south) |- (merge2);
\path [line] (merge2) -- (all_sharks.north);

\path [line] (all_sharks.east)
    -| node[above, pos=0.3] {Yes} (yes_all.north);

\path [line] (yes_all) -- (budget);

\path [line] (all_sharks.west)
    -- node[above] {No} ++(-5.8,0) |- (candidate.west);

\path [line] (budget.east)
    -- node[above] {No} ++(1.2,0) |- (progress.east);

\path [line] (budget.south) -- node[right] {Yes} (return.north);

\path [line] (return) -- (end);

\end{tikzpicture}
\caption{Workflow diagram of the MOSHO search cycle.}
\label{fig:workflow}
\end{figure}
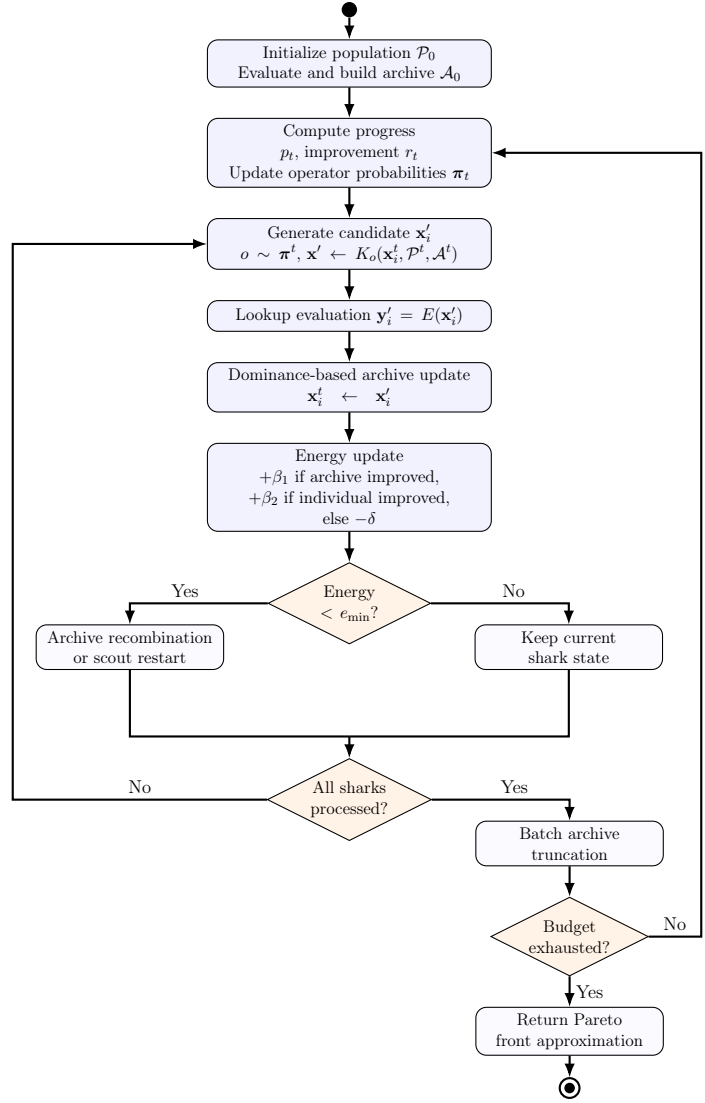

Algorithm~\ref{alg:mosho} gives the complete procedure and Figure~\ref{fig:workflow} the corresponding control flow. After initialization (lines~1--2), each iteration recomputes the two feedback signals and the operator probabilities (lines~4--5), then processes every shark in turn (lines~6--20): sample an operator, generate and evaluate a candidate, update the archive by dominance check, move the shark to the candidate if it improves the archive or its current solution, and regulate its energy according to Eq.~\eqref{eq:energy}, restarting a depleted agent when needed. Only once all agents have been processed is the archive truncated to capacity (line~21), which is the deferral analyzed in Section~\ref{subsec:truncation}. When the budget $B$ is exhausted, the final archive is returned as the Pareto-front approximation $\hat{P}$.

\begin{algorithm}[htbp]
\caption{MOSHO: Multi-Objective Shark Hunting Optimization}
\label{alg:mosho}
\footnotesize
\begin{algorithmic}[1]
\REQUIRE $N$, budget $B$, archive limit $M$, base probabilities $\boldsymbol{\pi}^0$, energy parameters $e_0, e_{\min}, e_{\max}, \beta_1, \beta_2, \delta, \gamma$
\ENSURE Pareto archive $\mathcal{A}$

\STATE Sample $\mathcal{P}^0 = \{(\mathbf{x}_i^0,e_0)\}_{i=1}^{N}$ uniformly over $\mathcal{X}$; evaluate; build $\mathcal{A}^0$
\STATE eval\_count $\leftarrow N$

\WHILE{eval\_count $< B$}
    \STATE $p^t \leftarrow \text{eval\_count}/B$;\quad
           $r^t \leftarrow$ archive improvement over window $H$
    \STATE $\boldsymbol{\pi}^t \leftarrow
           \mathrm{Normalize}(\boldsymbol{\pi}^0 \odot g(r^t,p^t))$

    \FOR{each shark $i = 1,\ldots,N$}
        \STATE \textbf{if} eval\_count $\geq B$ \textbf{then break}
        \STATE $o \sim \boldsymbol{\pi}^t$;\quad
               $\mathbf{x}' \leftarrow
               K_o(\mathbf{x}_i^t,\mathcal{P}^t,\mathcal{A}^t)$
        \STATE $\mathbf{y}' \leftarrow \mathcal{E}(\mathbf{x}')$;\quad
               eval\_count $\leftarrow$ eval\_count $+1$
        \STATE \texttt{arch\_imp} $\leftarrow$
               \textsc{UpdateArchive}$(\mathbf{x}',\mathcal{A}^t)$
               \COMMENT{$\mathcal{O}(A)$}
        \STATE \texttt{ind\_imp} $\leftarrow$
               $\left[\mathbf{y}' \text{ Pareto-dominates } \mathbf{y}_i^t\right]$

        \IF{\texttt{arch\_imp}}
            \STATE $\mathbf{x}_i^t \leftarrow \mathbf{x}'$;\quad
                   $\mathbf{y}_i^t \leftarrow \mathbf{y}'$;\quad
                   $e_i \leftarrow \min(e_{\max},e_i+\beta_1)$
        \ELSIF{\texttt{ind\_imp}}
            \STATE $\mathbf{x}_i^t \leftarrow \mathbf{x}'$;\quad
                   $\mathbf{y}_i^t \leftarrow \mathbf{y}'$;\quad
                   $e_i \leftarrow \min(e_{\max},e_i+\beta_2)$
        \ELSE
            \STATE $e_i \leftarrow \max(0,e_i-\delta)$
        \ENDIF

        \STATE \textbf{if} $e_i < e_{\min}$ \textbf{then}
               $\mathbf{x}_i^t \leftarrow
               \textsc{ArchiveRecomb}(\mathcal{A}^t)$ or
               $\textsc{Scout}()$;\quad
               $e_i \leftarrow \max(\gamma e_0,e_{\min})$
    \ENDFOR

    \STATE \textbf{if} $|\mathcal{A}^t| > M$ \textbf{then}
           $\textsc{CrowdingTruncate}(\mathcal{A}^t,M)$
           \COMMENT{$\mathcal{O}(A\log A)$, once per iteration}
\ENDWHILE

\RETURN $\mathcal{A}$
\end{algorithmic}
\end{algorithm}

\FloatBarrier
\section{Comprehensive Experimental Evaluation}
\label{sec:results}

This section presents an empirical evaluation of the proposed MOSHO algorithm across 18 dataset--device contexts derived from the merged NAS-Bench-201 \citep{dong2020nasbench201} and HW-NAS-Bench \citep{li2021hwnas} lookup tables. The evaluation assesses the accuracy--latency trade-offs achieved by the search mechanisms and analyzes the impact of the proposed discrete algorithmic components.

All experiments rely on deterministic tabular evaluation and are executed on a single Windows~11~Pro workstation equipped with an AMD Ryzen~7 5800H processor, 32~GB DDR4 RAM, and an NVIDIA RTX~3060 GPU with 6~GB VRAM. The evaluation benchmarks cover six distinct edge hardware devices (EdgeGPU, EdgeTPU, Eyeriss, FPGA, Pixel~3, and Raspberry~Pi~4) across three representative vision datasets: CIFAR-10 (validation split), CIFAR-100, and ImageNet16-120.
The evaluation protocol considers six evaluation budgets $B \in \{1000, 2000, 5000, 7000, 10000, 15000\}$, with 30 independent runs per method per context using deterministic random seeds defined by $s_n = 42 + (n-1)\times 7$, for $n=1,\ldots,30$. Statistical significance is assessed using paired Wilcoxon signed-rank tests \citep{wilcoxon1945} with Holm--Bonferroni adjustment, complemented by Vargha--Delaney $A_{12}$ effect sizes \citep{vargha2000}.

To ensure full reproducibility, the ORAMA-NAS framework which implements the proposed MOSHO algorithm, all baseline methods, and the complete experimental benchmarking pipeline is made publicly available at \href{https://github.com/Orama-team/orama-nas-framework}{https://github.com/Orama-team/orama-nas-framework}.

The remainder of this section is organized as follows: Section~\ref{subsec:metrics} defines the evaluation metrics; the baseline methods and hyperparameter configuration are then presented; Section~\ref{subsec:ablation} examines the contribution of each algorithmic component; Sections~\ref{subsec:performance} and~\ref{subsec:comparative} assess MOSHO's performance and compare it against baseline methods; Section~\ref{sec:sensitivity_analysis} investigates sensitivity to population size and evaluation budget; and Section~\ref{subsec:scalability} examines scalability across an expanded search space and a three-objective formulation.
\subsection{Evaluation Metrics}
\label{subsec:metrics}

All metrics are reported as medians over 30 independent runs per method per context. Table~\ref{tab:evaluation_metrics} summarizes the evaluated metrics, their primary interpretations, and optimization goals. Detailed mathematical formulations and reference point derivations are provided in \ref{app:metrics} of the Supplementary Materials.

\begin{table}[htbp]
\centering
\small
\caption{Summary of Evaluation Metrics}
\label{tab:evaluation_metrics}
\renewcommand{\arraystretch}{0.9}
\setlength{\tabcolsep}{3pt}
\begin{tabularx}{0.95\linewidth}{@{} l X c @{}}
\toprule
\textbf{Metric} & \textbf{Description} & \textbf{Goal} \\
\midrule
HV & Measures both convergence and diversity by computing the objective volume dominated by the approximation front $\hat{P}$ relative to a reference point $(r_1, r_2)$ offset by $\varepsilon = 10^{-9}$ from $P^*$ & $\uparrow$ \\
\midrule
nHV & Normalizes HV by the hypervolume of the exact Pareto front ($\mathrm{HV}(P)/\mathrm{HV}_{\max}$) to enable fair comparison across contexts and scales & $\uparrow$ \\
\midrule
$\text{IGD}^+$ & Evaluates both proximity and coverage by computing the average minimum modified distance from $P^*$ to $\hat{P}$ while preserving weak Pareto compliance & $\downarrow$ \\
\midrule
Spacing ($S$) & Evaluates the uniformity of distribution between adjacent non-dominated solutions via the standard deviation of nearest-neighbour distances $d_i$ & $\downarrow$ \\
\midrule
Execution Time & Measures computational efficiency and algorithmic overhead via median wall-clock runtime across complete search runs at each budget level & $\downarrow$ \\
\bottomrule
\multicolumn{3}{@{}c@{}}{\scriptsize Note: $\uparrow$ means maximize; $\downarrow$ means minimize.} \\
\end{tabularx}
\end{table}
\subsection{Baseline Methods}
MOSHO is compared against five methods:
\begin{itemize}[itemsep=0pt, topsep=2pt, parsep=0pt]
  \item \textbf{MOWSO} \citep{zheng2025}: MOSHO's direct predecessor; matched archive limit $M=50$ for fairness.
    \item \textbf{NSGA-II} \citep{kumar2025}: Canonical multi-objective evolutionary algorithm.
  \item \textbf{ABC (HiveNAS)} \citep{shahawy2023}: Artificial Bee Colony-based NAS with on-the-fly architecture encoding.
  \item \textbf{Firefly Algorithm} \citep{nguyen2025}: A rank-based multi-objective firefly algorithm adapted for NAS.
  \item \textbf{MOPSO} \citep{coello2004}: Multi-objective particle swarm optimization adapted for discrete NAS.
\end{itemize}
\subsection{Hyperparameter Configuration}

MOSHO hyperparameters were tuned via Optuna \citep{akiba2019optuna} using a Tree-structured Parzen Estimator (TPE) model on a held-out validation context. Table~\ref{tab:params} reports all settings.

\begin{table}[H]
\caption{MOSHO hyperparameter settings after Optuna tuning.}
\label{tab:params}
\centering
\footnotesize
\renewcommand{\arraystretch}{1.15}
\setlength{\tabcolsep}{4pt}
\resizebox{\textwidth}{!}{%
\begin{tabular}{@{}llcc@{\hspace{10pt}}|@{\hspace{10pt}}llcc@{}}
\toprule
\textbf{Param} & \textbf{Description} & \textbf{Range} & \textbf{Value} & \textbf{Param} & \textbf{Description} & \textbf{Range} & \textbf{Value} \\
\midrule
$N$        & Population size          & $\{20,30,50,80,100\}$ & 50    & $\gamma$            & Energy reset factor   & $[0,1]$ & 0.50 \\
$M$        & Archive capacity         & $[10, N]$              & 50    & $P_{\rm patrol}$    & Patrol probability    & $[0,1]$ & 0.20 \\
$e_0$      & Initial energy/agent     & $[0.5,\,2.0]$          & 0.818 & $P_{\rm scent}$     & Scent probability     & $[0,1]$ & 0.25 \\
$e_{\min}$ & Restart energy thresh.   & $[0.05,\,0.5]$         & 0.500 & $P_{\rm circle}$    & Circle probability    & $[0,1]$ & 0.15 \\
$e_{\max}$ & Energy cap               & $[0.5,\,2.0]$          & 1.699 & $P_{\rm burst}$     & Burst probability     & $[0,1]$ & 0.10 \\
$\delta$   & Energy decay (rejection) & $[0,\,0.1]$            & 0.0968 & $P_{\rm crossover}$ & Crossover probability & $[0,1]$ & 0.20 \\
$\beta_1$  & Energy gain (arch.\ improved) & $[0,1]$          & 0.20  & $P_{\rm scout}$     & Scout probability     & $[0,1]$ & 0.10 \\
$\beta_2$  & Energy gain (accept, no arch.) & $[0,1]$         & 0.05  & $H$                 & Improvement window (iter.) & -- & 5 \\
\bottomrule
\end{tabular}%
}
\end{table}

With the obtained tuned values ($\gamma=0.50$, $e_0=0.818$,
$e_{\min}=0.500$), $\gamma e_0=0.409<e_{\min}$. Thus, the reset rule
$\max(\gamma e_0,e_{\min})$ evaluates to $e_{\min}$ for the reported
configuration.

\subsection{Ablation Analysis}
\label{subsec:ablation}

To isolate the contribution of individual components, we evaluate all
single-unit removals and the grouped variants listed in
Table~\ref{tab:ablation_units}. MOSHO denotes the final adopted
configuration, while MOSHO-Full denotes the complete configuration
including all development components.

Three of the tested units (U02, U08, and U9) are auxiliary mechanisms evaluated during development but excluded from the final MOSHO configuration. U02 (explicit acceptance policy) replaces unconditional acceptance with a structured three-case rule: a candidate is accepted outright if it Pareto-dominates the agent's current solution, accepted if it is non-dominated but improves archive coverage, and otherwise accepted only with a probability that decays as the search progresses, favoring free exploration early and selective convergence later; rejected candidates incur an energy penalty on the owning agent. U08 (K-means archive leader/seed selection) was a development-stage
alternative to the final archive sampling rule. It partitions the
archive in objective space using K-means and samples clusters
inversely proportional to their size, thereby favoring sparse regions.
This mechanism was excluded from the final MOSHO configuration after
showing no consistent benefit. U9 (mono-objective fallback) monitors the spread of each objective across the archive and, if one objective's range collapses below a small threshold (a degeneracy that can occur on hardware where many architectures share identical latency), temporarily suspends Pareto-based comparison in favor of scalar comparison on the remaining objective, automatically reverting once the range recovers.

\begin{table}[H]
\caption{Ablation units and tested grouped variants.}
\label{tab:ablation_units}
\centering
\small
\renewcommand{\arraystretch}{1.15}
\setlength{\tabcolsep}{5pt}
\resizebox{\textwidth}{!}{%
\begin{tabular}{@{}llc@{\hspace{22pt}}llc@{}}
\toprule
\multicolumn{3}{c}{\textbf{Single-unit ablations}} & \multicolumn{3}{c}{\textbf{Grouped variants}} \\
\cmidrule(r){1-3}\cmidrule(l){4-6}
\textbf{Rem.} & \textbf{Name} & \textbf{Var.} & \textbf{Rem.} & \textbf{Name} & \textbf{Var.} \\
\midrule
U01 & Discrete operator kernels & ABL-U01 & U01+U02 & Search core removal & G-SEARCH \\
U02 & Explicit acceptance policy & ABL-U02 & U05+U06+U07 & Adaptive block removal & G-ADAPT \\
U03 & Energy-regulated restarts & ABL-U03 & U08 & K-means leader-selection removal & G-ARCHIVE \\
U04 & Dual adaptive signal & ABL-U04 & U08+U09 & No-advance mode & G-NOADV \\
U05 & Active archive feedback & ABL-U05 & U01--U07 & Core mechanism removal & G-NOBASE \\
U06 & Deferred truncation & ABL-U06 & U02+U08+U09 & Core archive/control removal & MOSHO \\
U07 & Restart source mixing & ABL-U07 & & & \\
U08 & K-means archive leader/seed selection & ABL-U08 & & & \\
U09 & No-advantage fallback & ABL-U09 & & & \\
\bottomrule
\end{tabular}%
}
\end{table}

\textbf{Statistical validation:} For each metric, we compare all the
MOSHO run series against every ablation variant using the paired Wilcoxon
signed-rank test~\citep{wilcoxon1945}. The resulting $p$-values are
adjusted with the Holm--Bonferroni procedure to control the family-wise
error rate across the ablation family. Effect sizes are reported using the
Vargha--Delaney $A_{12}$ statistic~\citep{vargha2000}; values above 0.50
favor MOSHO, values below 0.50 favor the ablated variant.

\begin{figure}[tp]
\centering
\includegraphics[width=0.85\linewidth]{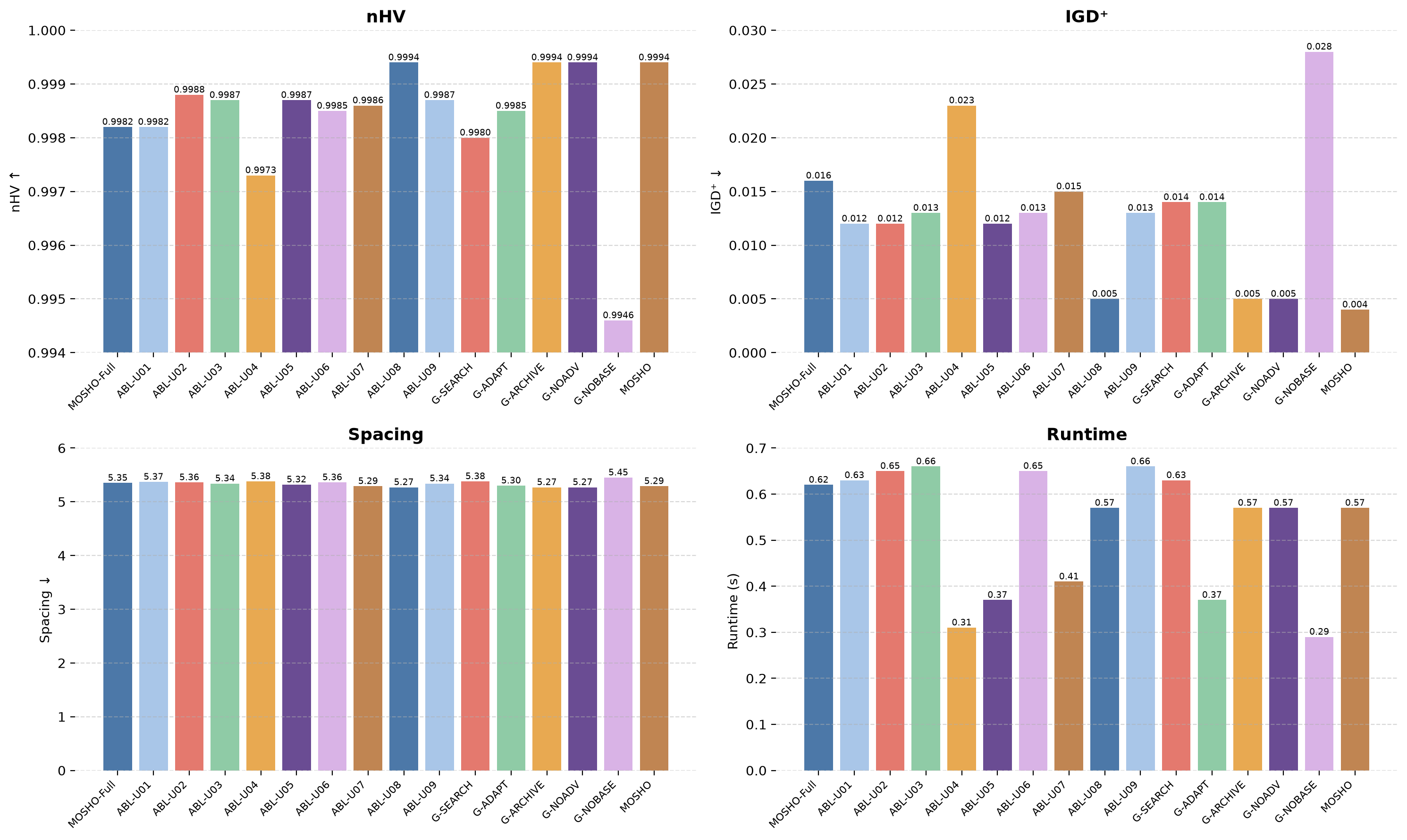}\\[2pt]
\includegraphics[width=0.85\linewidth]{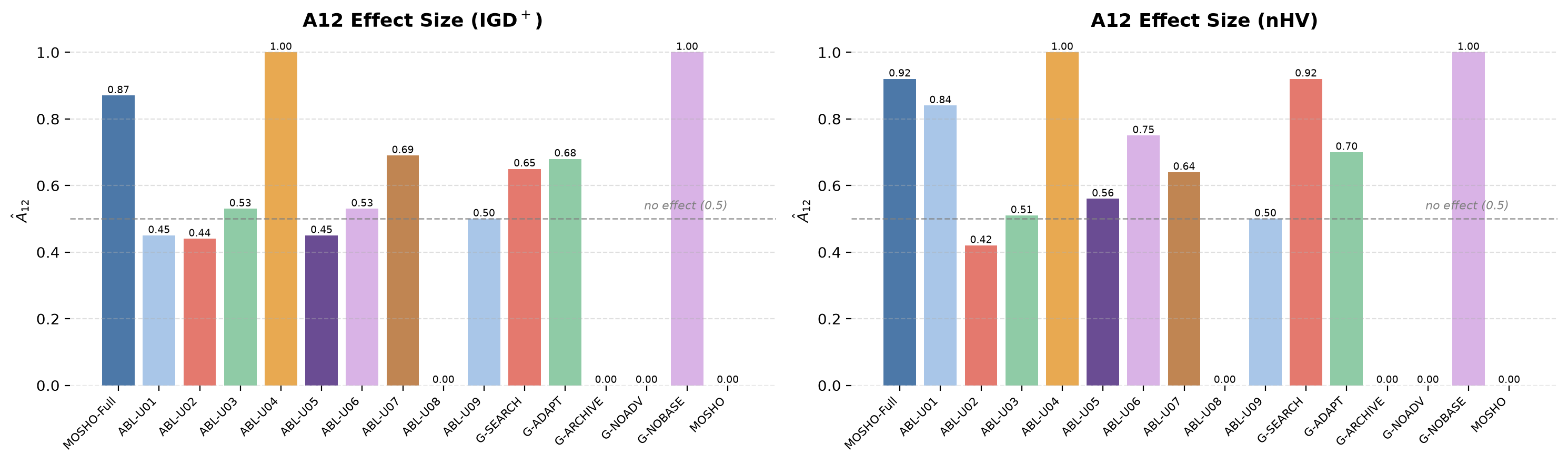}\\[2pt]
\includegraphics[width=0.85\linewidth]{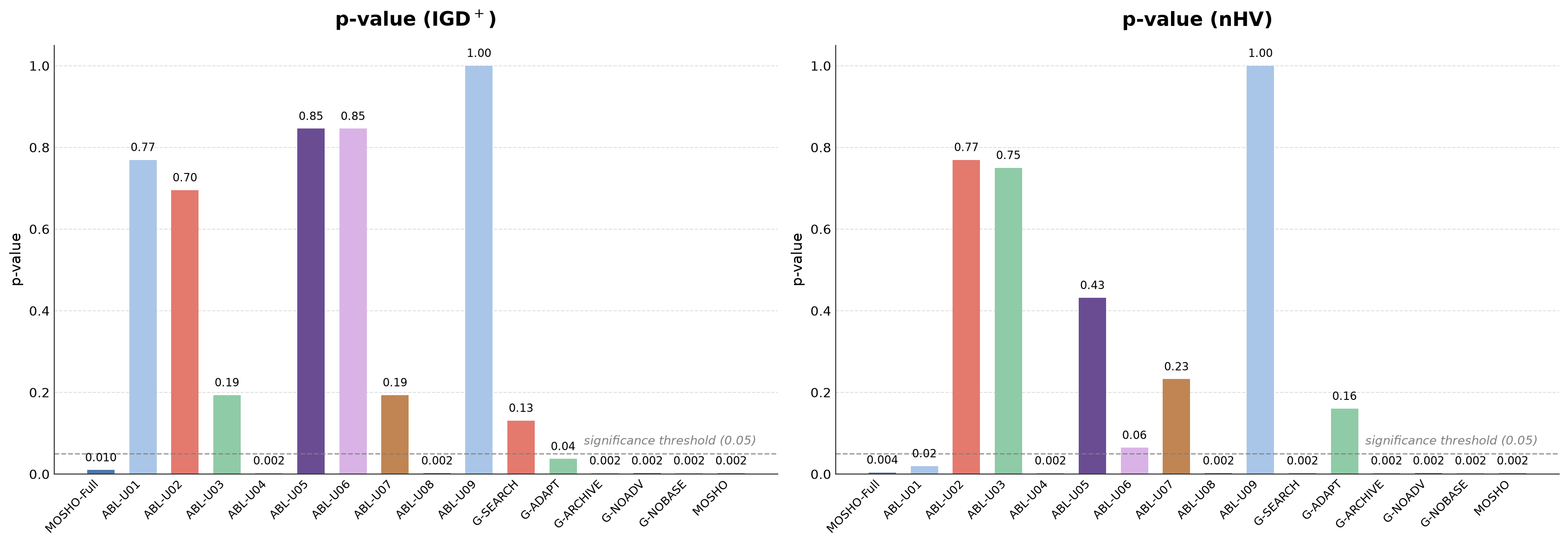}
\caption{Ablation study results across all tested variants. Top: median nHV, IGD$^{+}$, spacing, and execution time over 30 independent runs per variant across the 18 benchmark contexts. Middle: Vargha--Delaney $A_{12}$ effect sizes for IGD$^{+}$ (left) and nHV (right); values above 0.5 favor MOSHO, values below 0.5 favor the ablated variant, dashed line marks the no-effect threshold of 0.5. Bottom: Wilcoxon signed-rank $p$-values for IGD$^{+}$ (left) and nHV (right); values below the significance threshold ($\alpha=0.05$, dashed line) indicate a statistically significant difference from MOSHO.}
\label{fig:ablation_hv}
\end{figure}

Figure~\ref{fig:ablation_hv} reports the median nHV, IGD$^+$, spacing, and
execution time achieved by MOSHO and the tested ablation variants across the 18
benchmark contexts, together with the corresponding $A_{12}$ effect sizes
and the (unadjusted) Wilcoxon signed-rank $p$-values for IGD$^+$ and nHV.

\textbf{Key findings:}
\begin{itemize}
  \item The dual adaptive signal (U04) is the most influential single unit: removing it produces the largest single-unit deterioration in nHV and IGD$^{+}$, with complete stochastic dominance in MOSHO's favor ($A_{12}=1.0$ for both metrics).
  \item The development-stage K-means archive leader/seed mechanism (U08)
does not provide a consistent benefit on NAS-Bench-201-derived tasks and
is therefore excluded from the final MOSHO configuration.
  \item MOSHO, obtained by removing the development-only U02, U08, and U09
mechanisms, provides the strongest overall configuration in the ablation
study and is adopted as the final configuration used in all subsequent
experiments.
  \item G-NOBASE (core mechanisms removed) causes the largest overall degradation, confirming the mechanisms act synergistically rather than independently.
  \item U03, U09, U07 (energy restarts, mono-objective fallback, restart mixing) show limited individual impact ($A_{12}\approx0.50$), acting mainly through interaction with other components.
\end{itemize}

\subsection{Performance Assessment}
\label{subsec:performance}
Across the 18 dataset--device contexts, MOSHO achieves best, median, and
worst nHV values of 1.0000, 0.9987, and 0.9969, respectively. The
corresponding IGD$^{+}$ values are 0.0038, 0.0064, and 0.0067, while the
spacing values are 2.10, 2.74, and 3.55, respectively. The median execution time remained approximately constant at 0.64\,s across all contexts.

\textbf{Dataset-wise trends:} nHV is highest on CIFAR-10/100 and slightly lower on ImageNet16-120, consistent with its harder, more spread Pareto front; energy-regulated restarts prevent stagnation and sustain coverage through the late search phase.

\subsection{Comparative Study}
\label{subsec:comparative}

\begin{figure}[!htbp]
\centering
\includegraphics[width=0.85\linewidth]{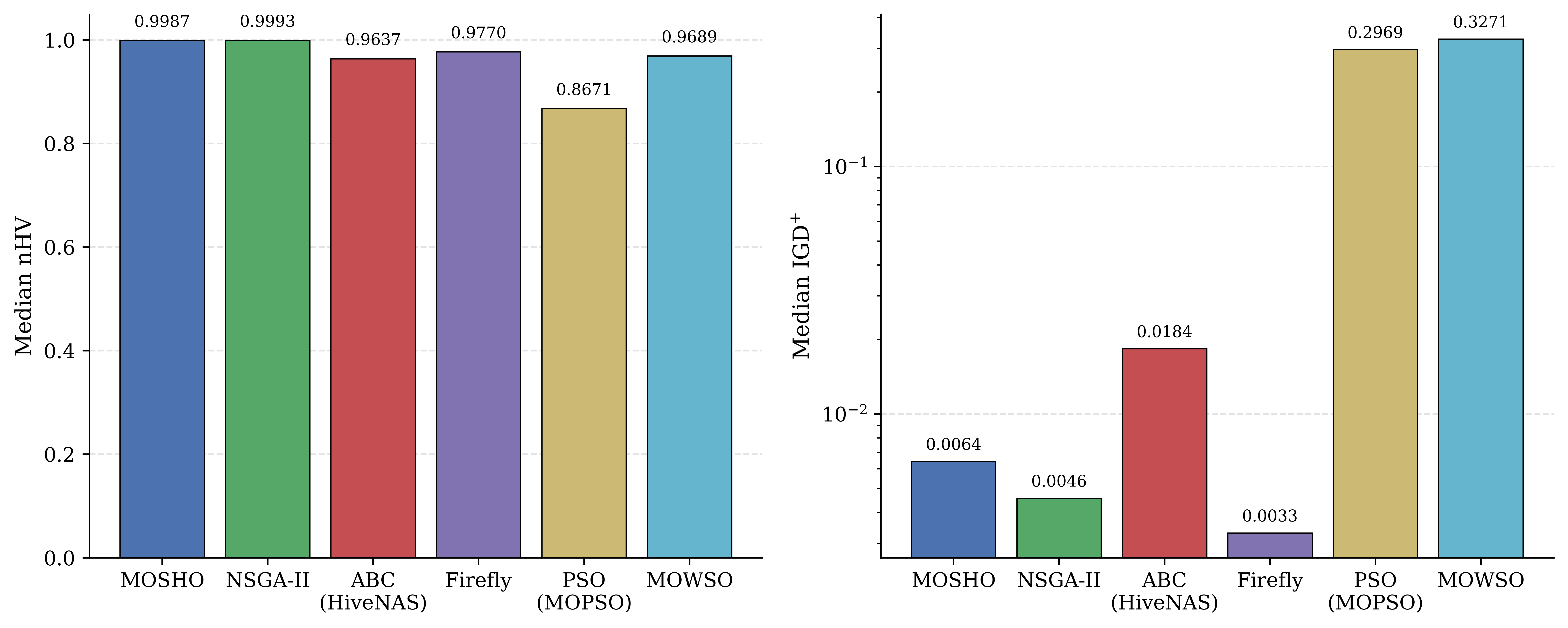}\\[-2pt]
\includegraphics[width=0.85\linewidth]{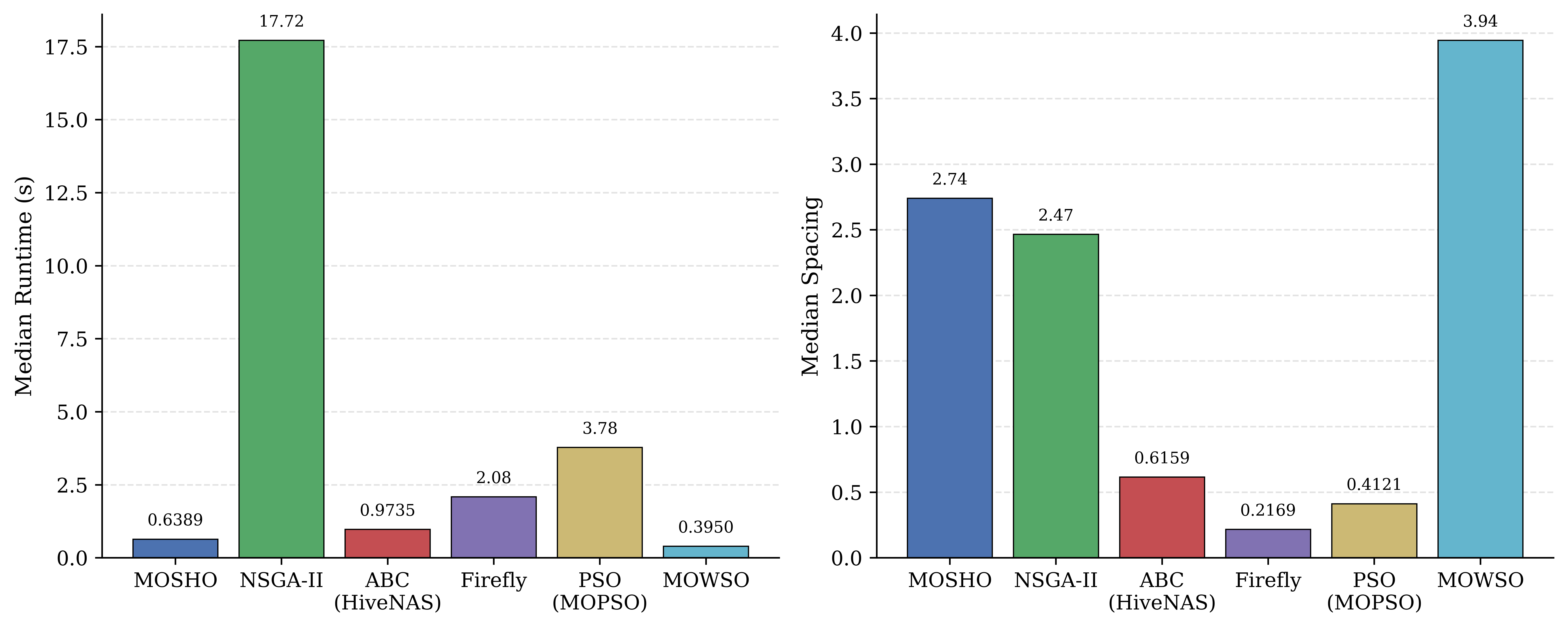}
\caption{Comparison of MOSHO with several multi-objective optimization algorithms across 18 contexts. Median nHV and IGD$^{+}$ (top), and execution time and spacing (bottom), are reported over 30 runs with a budget of 15000 evaluations.}
\label{fig:comparison_results}
\end{figure}

Figure~\ref{fig:comparison_results} compares all methods across the 18 contexts in terms of median nHV, IGD$^{+}$, spacing, and execution time. Results are aggregated over 30 runs with a budget of 15,000 evaluations.

\vspace{2em}
\textbf{Discussion}
\begin{itemize}
\item \textbf{Competitive convergence quality:}
MOSHO reaches a median nHV of 0.9987, close to the best method NSGA-II (0.9993), with a low median IGD$^{+}$ (0.0064). Firefly's lower IGD$^{+}$ (0.0033) comes at the cost of coverage (nHV 0.9770). MOSHO therefore provides a better balance between convergence and diversity, remaining highly competitive with NSGA-II while avoiding the coverage trade-off observed with Firefly.

\item \textbf{Computational efficiency:} MOSHO runs in a median of 0.64\,s, substantially faster than the
baselines while maintaining competitive solution quality; MOWSO's lower
runtime is accompanied by markedly weaker convergence and diversity.

\item \textbf{Solution diversity:} Firefly, MOPSO, and ABC show lower spacing but inferior convergence, confirming the usual diversity/accuracy trade-off; MOSHO's fronts remain well distributed.

\item \textbf{Overall:} MOSHO matches the strongest baselines closely on convergence and diversity at a fraction of their runtime, making it the practical choice when solution quality must be balanced against cost.
\end{itemize}

\subsection{Sensitivity Analysis}
\label{sec:sensitivity_analysis}

To assess robustness, we varied two of the most influential parameters, namely population size ($N$) and evaluation budget ($B$) independently, keeping all other hyperparameters at their Optuna-tuned values (Table~\ref{tab:params}) and following the protocol (18 contexts, 30 runs, median reported). Figures show per-metric normalized values; raw values are in \ref{app:supp_sensitivity_raw} of the Supplementary Materials.

\begin{figure}[H]
\centering
\begin{subfigure}{0.49\textwidth}
    \centering
    \includegraphics[width=\linewidth]{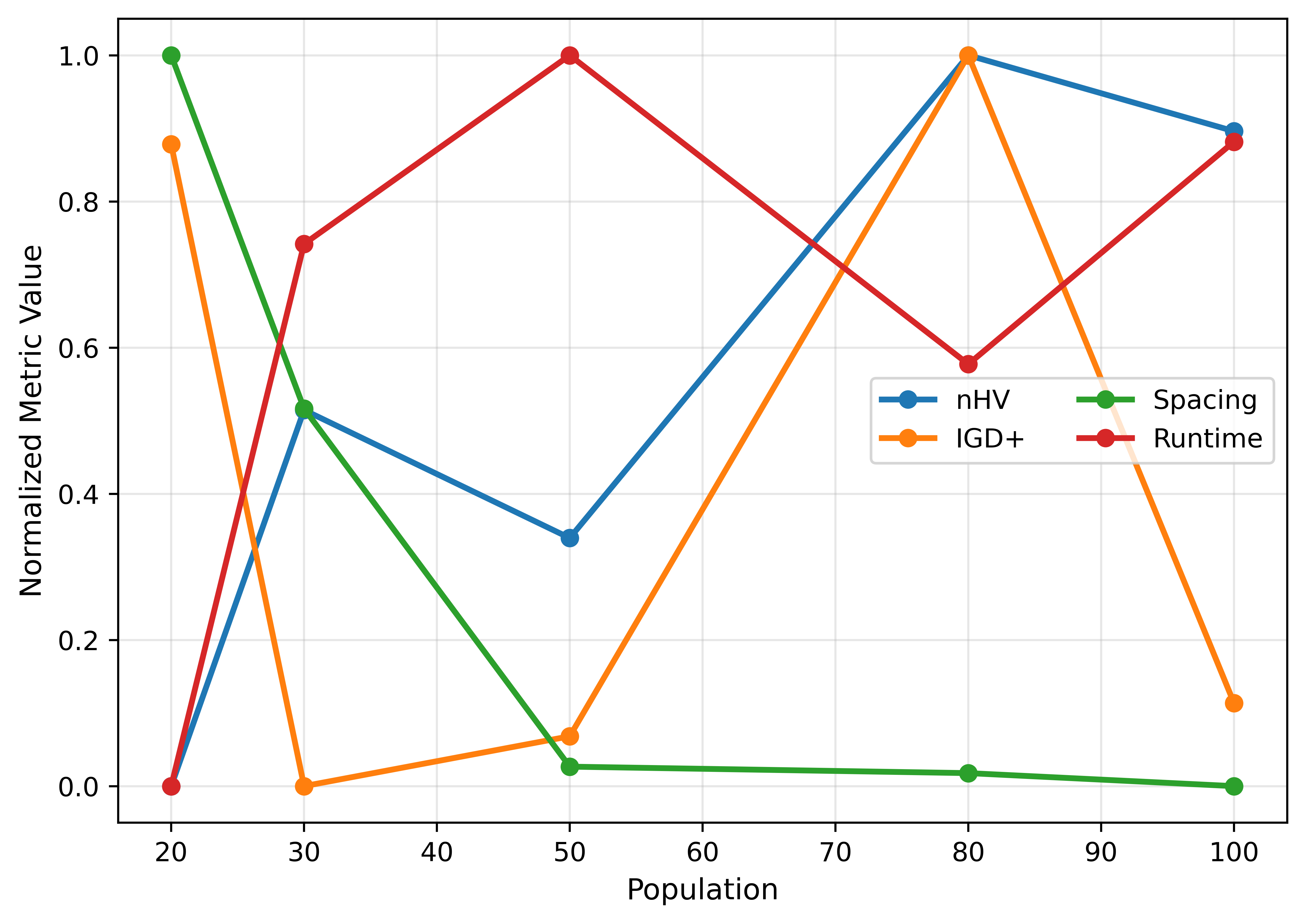}
    \caption{Effect of population size ($N$).}
    \label{fig:sensitivity_population}
\end{subfigure}\hfill
\begin{subfigure}{0.49\textwidth}
    \centering
    \includegraphics[width=\linewidth]{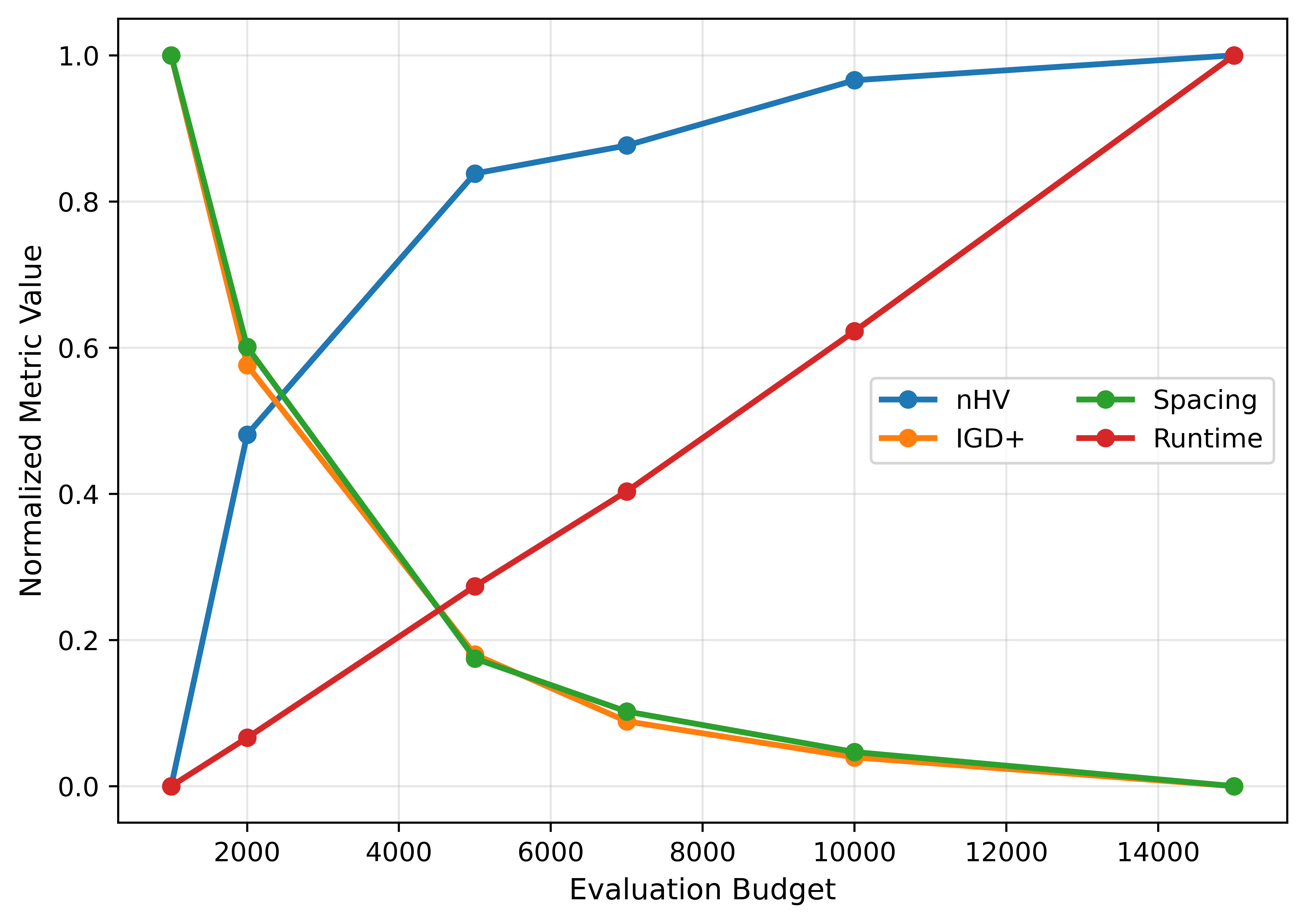}
    \caption{Effect of evaluation budget ($B$).}
    \label{fig:sensitivity_budget}
\end{subfigure}
\caption{Sensitivity analysis of MOSHO with respect to the two primary
hyperparameters. For visualization purposes, each metric is independently
normalized to the interval $[0,1]$, allowing the relative evolution of
nHV, IGD$^+$, spacing, and runtime to be compared on a common scale.}
\label{fig:sensitivity}
\end{figure}

The sensitivity analysis indicates that MOSHO is sensitive to its primary hyperparameters, achieving its best performance within the parameter ranges reported in Table~\ref{tab:params}.

\subsection{Scalability and Efficiency}
\label{subsec:scalability}

\begin{figure}[H]
    \centering
    \includegraphics[width=0.9\linewidth]{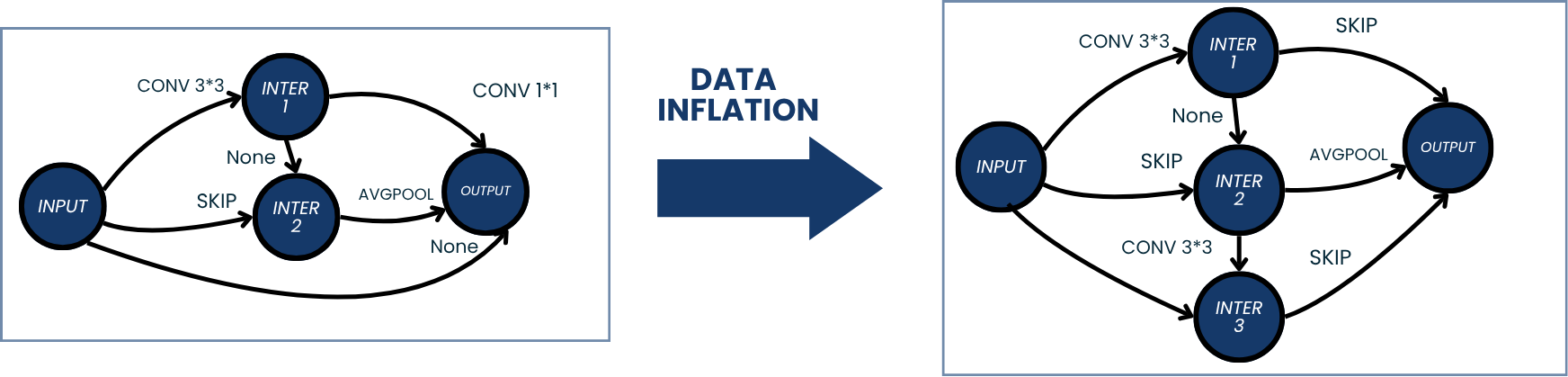}
    \caption{Data inflation procedure applied to the NAS-Bench-201 search
    space. A new intermediate node is inserted into the DAG, adding 4 new
    edges and expanding the architecture space from $5^{6}=15{,}625$ to
    $5^{10}=9{,}765{,}625$ candidate architectures.}
    \label{fig:scala}
\end{figure}

\textbf{Problem-size scalability:} The original NAS-Bench-201 space
($5^6 = 15{,}625$ architectures) was extended by adding one DAG node,
creating 4 new possible connections and expanding the space to
$5^{10} = 9{,}765{,}625$ architectures, as illustrated in
Figure~\ref{fig:scala}. Since exact evaluation is intractable at this
scale, a LightGBM surrogate model trained on NAS-Bench-201 observations
was used to predict accuracy and latency. MOSHO maintained stable
convergence on this larger space, with nHV degrading by less than 2\%
relative to the original benchmark.

\textbf{Objective-space scalability.} A three-objective NAS problem incorporating accuracy, latency, and energy consumption was evaluated against NSGA-III \citep{deb2014nsga3} and MOEA/D \citep{zhang2007}. This problem directly extends the bi-objective formulation of Section~\ref{subsec:formulation}: each architecture $\mathbf{x}$ is now characterized by three objective functions,
\begin{align}
  f_1(\mathbf{x}) &= \mathrm{Accuracy}(\mathbf{x}, \mathcal{D}) \quad \text{(maximize)} \\
  f_2(\mathbf{x}) &= \mathrm{Latency}_d(\mathbf{x}) \quad \text{(minimize)} \\
  f_3(\mathbf{x}) &= \mathrm{Energy}_d(\mathbf{x}) \quad \text{(minimize)}
\end{align}
where $f_3$ is obtained directly from the HW-NAS-Bench energy measurements for each candidate architecture. Pareto dominance, as defined in Section~\ref{subsec:formulation}, extends
naturally to this setting: a solution $\mathbf{a}$ dominates a solution $\mathbf{b}$, denoted $\mathbf{a} \prec \mathbf{b}$, if and only if $f_1(\mathbf{a}) \geq f_1(\mathbf{b})$, $f_2(\mathbf{a}) \leq f_2(\mathbf{b})$, and $f_3(\mathbf{a}) \leq f_3(\mathbf{b})$, with at least one inequality strict.

Under this explicit extension, the dual-signal scheduler continues to
dynamically adjust operator probabilities based on archive improvement
in the enlarged objective space, with no change required to the
scheduling logic itself. Across the 18 contexts, MOSHO consistently
outperforms NSGA-III and MOEA/D on all four metrics, as summarized in
Table~\ref{tab:mosho_results}. These results indicate that the archive-management and adaptive-scheduling
mechanisms remain effective when the objective space is extended to three
objectives.
\FloatBarrier
\begin{table}[htbp]
\centering
\caption{Comparison of MOSHO against NSGA-III and MOEA/D across 18 contexts.}
\label{tab:mosho_results}
\begin{tabular}{lccc}
\hline
\textbf{Metric} & \textbf{MOSHO} & \textbf{NSGA-III} & \textbf{MOEA/D} \\
\hline
nHV                & 0.9658 & 0.9350 & 0.9070 \\
IGD$^{+}$          & 0.042  & 0.063  & 0.066  \\
Spacing            & 0.628  & 4.248  & 7.702  \\
Runtime (s)        & 0.68   & 2.23  & 0.80   \\
\hline
\end{tabular}
\end{table}

\textbf{Computational efficiency and environmental cost:}
At a budget of 15,000 evaluations, MOSHO is the fastest at $\sim$0.64\,s,
compared with 17.72\,s for NSGA-II and 3.78\,s for MOPSO. Its energy
consumption (0.00052\,kWh) and CO$_2$ emissions (0.000311\,kg) remain
comparable to NSGA-II (0.00048\,kWh, 0.000306\,kg) and MOPSO
(0.00045\,kWh, 0.000280\,kg).

\section{Conclusions and Future Work}
\label{sec:conclusion}

This paper presented MOSHO, a native discrete multi-objective metaheuristic
for hardware-aware neural architecture search. MOSHO addresses the
limitations of continuous shark-inspired optimization in discrete
architecture spaces by replacing velocity--position dynamics and
post-hoc discretization with native discrete architecture-editing
operators. The resulting framework combines discrete search operators,
energy-regulated restarts, deferred archive truncation, and dual-signal
adaptive scheduling to balance exploration and exploitation efficiently.

Experimental evaluation on the merged NAS-Bench-201/HW-NAS-Bench benchmark
across 18 dataset--device contexts demonstrates that MOSHO achieves highly
competitive Pareto-front approximations with substantially lower runtime.
At 15,000 evaluations, MOSHO achieves a median nHV of 0.9987 and IGD$^{+}$
of 0.0064, with a median runtime of 0.64\,s, corresponding to a 96.4\%
reduction compared with NSGA-II (17.72\,s). Although NSGA-II achieves
slightly stronger median Pareto-quality metrics, MOSHO provides a favorable
trade-off between solution quality and computational cost. Scalability
experiments further show that MOSHO remains effective in an expanded
$5^{10}$ search space and when extending the problem from two to three
objectives.

Despite these encouraging results, the current evaluation has several
limitations. The experiments rely primarily on lookup-table benchmarks,
which enable deterministic and reproducible comparisons but do not capture
the full cost and variability of training architectures from scratch.
Moreover, the evaluation is centered on NAS-Bench-201-derived search spaces
and a fixed set of hardware targets. Further validation is therefore
required on larger search spaces, training-based NAS, and real hardware
platforms.

Future work will focus on extending MOSHO to larger and more diverse NAS
settings, including super-network-based and training-based search spaces.
We also plan to evaluate the method on real embedded hardware and broader
accelerator platforms, incorporating practical constraints such as
measurement variability and deployment cost. Finally, extending MOSHO to
many-objective HW-NAS by incorporating objectives such as memory footprint,
model size, and power consumption represents a promising direction.

\section*{Acknowledgments}

The authors thank Prof.\ Fatima Benbouzid-Si Tayeb and Prof.\ Malika Bessedik for their supervision throughout this project, and Dr.\ Hadjer Benmeziane for feedback on our work.

\bibliographystyle{elsarticle-harv}

\clearpage

\clearpage

\begin{center}
  {\LARGE\bfseries Supplementary Materials}
\end{center}

\appendix

\section{Mathematical Formulation of the Ancestor Algorithms}
\label{app:supp_ancestors}

This appendix recalls the update rules of the three shark-inspired optimizers from which MOSHO descends, in their original continuous form.

\subsection{White Shark Optimizer (WSO)}
\label{app:wso}

WSO \citep{braik2022} updates the velocity and position of search agents in a $D$-dimensional continuous space. The velocity of the $i$-th shark is
\begin{equation}
  v_i^{t+1} = \mu \left[ v_i^t + p_1 (x_{gbest}^t - x_i^t) \cdot c_1 + p_2 (x_{best, i}^t - x_i^t) \cdot c_2 \right],
  \label{eq:wso_velocity}
\end{equation}
where $x_{gbest}^t$ is the global best position, $x_{best,i}^t$ the personal best of agent $i$, and $c_1, c_2 \sim \mathcal{U}(0,1)$. The constriction factor $\mu$ modulates step sizes,
\begin{equation}
  \mu = \frac{2}{\left| 2 - \phi - \sqrt{\phi^2 - 4\phi} \right|}, \qquad \phi = p_1 + p_2 ,
  \label{eq:wso_constriction}
\end{equation}
and the acceleration coefficients follow fixed exponential schedules,
\begin{equation}
  p_1 = p_{\max} + (p_{\max} - p_{\min}) \cdot e^{-\left(\frac{4t}{T}\right)^2}, \quad
  p_2 = p_{\min} + (p_{\max} - p_{\min}) \cdot e^{-\left(\frac{4t}{T}\right)^2},
  \label{eq:wso_p_update}
\end{equation}
with $T$ the maximum number of iterations, $p_{\max} = 1.5$ and $p_{\min} = 0.5$. Positions are then updated according to the sensory intensity,
\begin{equation}
  x_i^{t+1} = \begin{cases}
    x_i^t + v_i^{t+1} \cdot f^{-1} & \text{if } r < mv_t, \\
    x_{gbest}^t + r_1 \cdot \vec{x}_w \cdot \operatorname{sgn}(r_2 - 0.5) & \text{if } r \geq mv_t,
  \end{cases}
  \label{eq:wso_position}
\end{equation}
where $r, r_1, r_2 \sim \mathcal{U}(0,1)$, $\operatorname{sgn}(r_2 - 0.5)$ allows direction changes, and $\vec{x}_w = |x_{gbest}^t - x_i^t|$. The wave frequency $f$ and the motion force $mv_t$ are
\begin{equation}
  f = f_{\min} + (f_{\max} - f_{\min}) \cdot \frac{1}{1 + e^{K_{\max} \cdot \left(\frac{t}{T} - a\right)}},
  \qquad
  mv_t = \frac{1}{a_0 + e^{a_1 \cdot \left(\frac{t}{T} - b\right)}},
  \label{eq:wso_freq_mv}
\end{equation}
with $f_{\min} = 0.07$, $f_{\max} = 0.75$, $a = 0.5$, $K_{\max} = T$, and $a_0 = a_1 = 1$, $b = 0.5$.

\subsection{Multi-Objective White Shark Optimizer (MOWSO)}
\label{app:mowso}

MOWSO \citep{zheng2025} replaces the global best by a leader drawn from the Pareto archive $\mathcal{A}^t$:
\begin{equation}
  v_i^{t+1} = \mu \left[ v_i^t + p_1 (x_{leader}^t - x_i^t) \cdot c_1 + p_2 (x_{best, i}^t - x_i^t) \cdot c_2 \right],
  \label{eq:mowso_velocity}
\end{equation}
where $x_{leader}^t \in \mathcal{A}^t$ is the archive member maximizing isolation. Archive capacity is enforced through the True Distance metric,
\begin{equation}
  TD(a) = \sum_{b \in \mathcal{A}^t \setminus \{a\}} \sqrt{\sum_{j=1}^{K} \left( \frac{f_j(a) - f_j(b)}{f_j^{\max} - f_j^{\min}} \right)^2},
  \label{eq:mowso_td}
\end{equation}
with $K$ the number of objectives and $f_j^{\max}, f_j^{\min}$ the current archive bounds of objective $j$. The member with the smallest $TD(a)$ is removed iteratively until $|\mathcal{A}^t| = M$.

\subsection{Shark Smell Optimization (SSO)}
\label{app:sso}

SSO \citep{kamarzaman2021} follows the gradient of the odor concentration. The forward-movement velocity and the resulting position are
\begin{equation}
  v_i^{t+1} = \eta_{t+1} \cdot R_1 \cdot \nabla(OF)\left|_{x_i^t}\right. + \alpha_{t+1} \cdot R_2 \cdot v_i^t,
  \qquad
  Y_i^{t+1} = x_i^t + v_i^{t+1} \cdot \Delta t,
  \label{eq:sso_forward}
\end{equation}
where $\eta_{t+1}$ is the gradient coefficient, $\alpha_{t+1}$ the momentum coefficient, $R_1, R_2 \sim \mathcal{U}(0,1)$, and $\Delta t$ the time step. When forward movement fails to increase the odor concentration, a rotational local search generates $M_{ls}$ candidates,
\begin{equation}
  Z_{i, m}^{t+1} = Y_i^{t+1} + R_3 \cdot Y_i^{t+1}, \quad m = 1, \dots, M_{ls}, \quad R_3 \sim \mathcal{U}(-1,1),
  \label{eq:sso_rotational}
\end{equation}
and the next position is the best of the forward and rotational candidates,
\begin{equation}
  x_i^{t+1} = \arg\max \left\{ OF(Y_i^{t+1}), OF(Z_{i, 1}^{t+1}), \dots, OF(Z_{i, M_{ls}}^{t+1}) \right\}.
  \label{eq:sso_selection}
\end{equation}

\section{Structural Comparison between MOSHO and its Ancestor Methods}
\label{app:supp_comparison}

Table~\ref{tab:comparison} contrasts MOSHO with WSO \citep{braik2022}, MOWSO \citep{zheng2025}, and SSO \citep{kamarzaman2021} along the dimensions that matter for discrete hardware-aware NAS. $N$ denotes the population size and $A$ the current archive size.

\begin{table}[H]
\caption{Structural comparison between MOSHO and its ancestor methods.}
\label{tab:comparison}
\centering
\scriptsize
\setlength{\tabcolsep}{3pt}
\renewcommand{\arraystretch}{1.05}
\begin{tabularx}{\textwidth}{@{}>{\raggedright\arraybackslash}p{1.85cm} >{\hsize=0.72\hsize\raggedright\arraybackslash}X >{\hsize=1.16\hsize\raggedright\arraybackslash}X >{\hsize=0.76\hsize\raggedright\arraybackslash}X >{\hsize=1.36\hsize\raggedright\arraybackslash}X@{}}
\toprule
\textbf{Dimension} & \textbf{WSO} & \textbf{MOWSO} & \textbf{SSO} & \textbf{MOSHO} \\
\midrule
Objective type & Single & Multi & Single & Multi \\
Search space & Continuous & Continuous & Continuous & Discrete (native) \\
Search mechanism & Velocity--position, factor $\mu$ & Velocity--position with post-hoc rounding & Gradient-following with rotational search & Six discrete operator kernels \\
Crossover & None & None & None & Uniform, then mutation ($p\!=\!0.3$) \\
Archive strategy & None & Pareto archive, True Distance & None & Pareto archive, crowding distance \\
Truncation timing & N/A & Per candidate & N/A & End of iteration (batch) \\
Truncation cost & N/A & $\mathcal{O}(N \!\cdot\! A^3)$ & N/A & $\mathcal{O}(N \!\cdot\! A \!+\! A\!\log\! A)$ \\
Restart & None & None & None & Energy-regulated, archive recombination \\
Adaptation & Fixed schedules & Fixed exponential decay & Adaptive (gradient) & Dual signal $r^t$, $p^t$ \\
Archive feedback & N/A & Passive (leader only) & N/A & Active (operators, restarts) \\
Boundary preservation & N/A & Not guaranteed & N/A & Extremes excluded from pruning \\
\bottomrule
\end{tabularx}
\end{table}

Three differences are decisive. First, MOSHO operates natively in a discrete space, with no continuous relaxation or post-hoc discretization, avoiding the quantization step required by MOWSO for integer-encoded NAS. Second, MOSHO combines energy-regulated restarts with archive-guided recombination and adaptive operator selection, providing explicit mechanisms for maintaining exploration when the search stagnates. Third, deferred batch truncation replaces repeated archive pruning with a single crowding-distance operation per iteration, reducing the archive-management cost from the $\mathcal{O}(N\!\cdot\!A^3)$ per-iteration cost of MOWSO to $\mathcal{O}(N\!\cdot\!A + A\log A)$.

\section{Adaptive Operator Scheduling}
\label{app:supp_extensions}

The operator-specific feedback function $g(r^t,p^t)$ used in the probability update $\boldsymbol{\pi}^t = \mathrm{Normalize}\bigl(\boldsymbol{\pi}^0 \odot g(r^t, p^t)\bigr)$ of the main text is defined piecewise as follows, where $r^t \in [0,1]$ is the archive improvement rate over the last $H = 5$ iterations and $p^t \in [0,1]$ the consumed fraction of the evaluation budget:
\begin{equation}
g(r^t,p^t)=
\begin{cases}
1+0.6(1-p^t), & \text{Patrol},\\
1+0.6p^t, & \text{Scent},\\
1+0.25r^t, & \text{Circle},\\
1+0.3\max(0,0.2-r^t), & \text{Burst},\\
1+0.5p^t, & \text{Crossover},\\
1+(1-p^t)(1-r^t), & \text{Scout}.
\end{cases}
\label{eq:feedback_function}
\end{equation}

\section{Mathematical Formulation of Evaluation Metrics}
\label{app:metrics}

This appendix provides the formal mathematical formulations of the multi-objective performance metrics used in the evaluation.

\subsection{Hypervolume (HV)}
\label{app:metric_hv}
The Hypervolume metric measures the volume of the objective space dominated by the Pareto front approximation $\hat{P}$:
\begin{align}
  \mathrm{HV}(P)
  &= \lambda\!\left(
  \bigcup_{x \in P}
  [r_1,f_1(x)] \times [f_2(x),r_2]
  \right), \label{eq:hv}\\
  r_1 &= \min_{p \in P^*} f_1(p) - \varepsilon, \qquad
  r_2 = \max_{p \in P^*} f_2(p) + \varepsilon, \notag
\end{align}
where $\lambda$ denotes the Lebesgue measure, $(r_1,r_2)$ is the context-specific reference point obtained from the exact Pareto front $P^*$, and $\varepsilon=10^{-9}$ (accuracy maximized, latency minimized), guaranteeing that every point of $P^*$ strictly dominates $r$. The same reference point is used across all methods for comparability.

\subsection{Normalized Hypervolume (nHV)}
\label{app:metric_nhv}
The Normalized Hypervolume scales the hypervolume relative to that of the exact Pareto front for each context:
\begin{equation}
  \mathrm{nHV} = \frac{\mathrm{HV}(P)}{\mathrm{HV}_{\max}}, \label{eq:nhv}
\end{equation}
where $\mathrm{HV}_{\max} = \mathrm{HV}(P^*)$. Higher values indicate superior front approximation.

\subsection{Modified Inverted Generational Distance ($\mathrm{IGD}^+$)}
\label{app:metric_igd}
$\mathrm{IGD}^+$ computes the average, over $P^*$, of the minimum modified distance to the approximation front $\hat{P}$:
\begin{equation}
  \mathrm{IGD}^+(P, P^*) = \frac{1}{|P^*|} \sum_{y \in P^*} \min_{x \in P} d^+(x, y). \label{eq:igdplus}
\end{equation}
Lower values are better, rewarding both proximity and coverage.

\subsection{Spacing ($S$)}
\label{app:metric_spacing}
Spacing measures the standard deviation of nearest-neighbour distances in the reported objective representation:
\begin{equation}
  S = \sqrt{\frac{1}{|P|-1}\sum_{i=1}^{|P|}(\bar{d} - d_i)^2}, \quad d_i = \min_{j \neq i} \|f_i - f_j\|, \label{eq:spacing}
\end{equation}
where $d_i$ is solution $i$'s distance to its nearest neighbour and $\bar{d}$ is the mean of all $d_i$. Lower values indicate a more uniform distribution.

\subsection{Execution Time}
\label{app:metric_time}
Execution time is evaluated as the median wall-clock time in seconds for one complete run across 30 independent runs at each evaluation budget level.

\section{Raw Sensitivity Analysis Values}
\label{app:supp_sensitivity_raw}

Tables~\ref{tab:sensitivity_budget_raw} and~\ref{tab:sensitivity_population_raw} report the raw (unnormalized) median values for the evaluation budget $B$ and the population size $N$, respectively.

\begin{table}[H]
\caption{Effect of evaluation budget $B$ on raw median metrics ($N=50$ fixed).}
\label{tab:sensitivity_budget_raw}
\centering
\small
\renewcommand{\arraystretch}{1.2}
\begin{tabular}{cccccc}
\toprule
$B$ & HV & IGD$^{+}$ & Spacing & Runtime (s) & Best Latency \\
\midrule
1{,}000  & 0.9881 & 0.1120 & 3.2454 & 0.0538 & 0.3571 \\
2{,}000  & 0.9935 & 0.0674 & 3.0531 & 0.1150 & 0.3568 \\
5{,}000  & 0.9974 & 0.0258 & 2.8476 & 0.3066 & 0.3567 \\
7{,}000  & 0.9979 & 0.0162 & 2.8126 & 0.4265 & 0.3566 \\
10{,}000 & 0.9989 & 0.0110 & 2.7859 & 0.6289 & 0.3567 \\
15{,}000 & 0.9992 & 0.0069 & 2.7634 & 0.9779 & 0.3566 \\
\bottomrule
\end{tabular}
\end{table}

\begin{table}[H]
\caption{Effect of population size $N$ on raw median metrics ($B=15{,}000$ fixed).}
\label{tab:sensitivity_population_raw}
\centering
\small
\renewcommand{\arraystretch}{1.2}
\begin{tabular}{cccccc}
\toprule
$N$ & HV & IGD$^{+}$ & Spacing & Runtime (s) & Best Latency \\
\midrule
20  & 0.9992 & 0.0089 & 3.8056 & 0.8921 & 0.3565 \\
30  & 0.9993 & 0.0067 & 3.2879 & 0.9756 & 0.3566 \\
50  & 0.9992 & 0.0069 & 2.7634 & 1.0047 & 0.3566 \\
80  & 0.9994 & 0.0092 & 2.7539 & 0.9571 & 0.3566 \\
100 & 0.9993 & 0.0070 & 2.7348 & 0.9914 & 0.3566 \\
\bottomrule
\end{tabular}
\end{table}

\end{document}